\pdfoutput=1

\documentclass[11pt]{article}

\usepackage[]{acl}

\usepackage{times}
\usepackage{latexsym}

\usepackage[T1]{fontenc}

\usepackage[utf8]{inputenc}

\usepackage{microtype}

\usepackage{inconsolata}

\usepackage{hyperref}
\usepackage{url}
\usepackage{microtype}
\usepackage{amsfonts}
\usepackage{mathrsfs}
\usepackage{amsmath}
\usepackage{listings}
\usepackage{subcaption}
\usepackage{multirow} 
\usepackage{multicol}
\usepackage{graphicx}
\usepackage{tikz}
\usepackage{xcolor}
\usepackage{pgfplots}
\usepackage{hyperref}
\usepgfplotslibrary{groupplots}
\usepackage{adjustbox}
\usepackage{wrapfig}
\usepackage{rotating}
\pgfplotsset{compat=1.16,
        /pgfplots/ybar legend/.style={
        /pgfplots/legend image code/.code={%
        \draw[##1,/tikz/.cd,bar width=3pt,yshift=-0.2em,bar shift=0pt]
                plot coordinates {(0cm,0.8em)};},
},
}
\definecolor{darkyellow}{rgb}{0.9, 0.7, 0.0}
\definecolor{bblue}{HTML}{4F81BD}
\definecolor{rred}{HTML}{C0504D}
\definecolor{ggreen}{HTML}{9BBB59}
\definecolor{ppurple}{HTML}{9F4C7C}
\newcommand{\mycases}[1]{\small\texttt{#1}}

\title{From Language Modeling to Instruction Following:\\ Understanding the Behavior Shift in LLMs after Instruction Tuning}

\author{Xuansheng Wu$^{\clubsuit}$\thanks{Work done during the internship at Tencent AI Lab.}\thanks{Contact Email: xuansheng.wu@uga.edu.}, Wenlin Yao$^\heartsuit$\thanks{Contact Email: wenlinyao@global.tencent.com.}, Jianshu Chen$^\heartsuit$, \\
\textbf{Xiaoman Pan$^\heartsuit$, Xiaoyang Wang$^\heartsuit$, Ninghao Liu$^\clubsuit$, Dong Yu$^\heartsuit$} \\
$^\clubsuit$University of Georgia\,\,\,\,$^\heartsuit$Tencent AI Lab, Bellevue
}

\begin{document}
\maketitle

\begin{abstract}
Large Language Models (LLMs) have achieved remarkable success, where instruction tuning is the critical step in aligning LLMs with user intentions. 
In this work, we investigate how the instruction tuning adjusts pre-trained models with a focus on intrinsic changes. 
Specifically, we first develop several local and global explanation methods, including a gradient-based method for input-output attribution, and techniques for interpreting patterns and concepts in self-attention and feed-forward layers. 
The impact of instruction tuning is then studied by comparing the explanations derived from the pre-trained and instruction-tuned models.
This approach provides an internal perspective of the model shifts on a human-comprehensible level.
Our findings reveal three significant impacts of instruction tuning: 
1) It empowers LLMs to recognize the instruction parts of user prompts, and promotes the response generation constantly conditioned on the instructions. 
2) It encourages the self-attention heads to capture more word-word relationships about instruction verbs.
3) It encourages the feed-forward networks to rotate their pre-trained knowledge toward user-oriented tasks.  
These insights contribute to a more comprehensive understanding of instruction tuning and lay the groundwork for future work that aims at explaining and optimizing LLMs for various applications. 
Our code and data are publicly available at \url{https://github.com/JacksonWuxs/Interpret_Instruction_Tuning_LLMs}.
\end{abstract}

\section{Introduction}
The strong capability of Large Language Models (LLMs) to align with user intentions is well-recognized across various real-world applications, where they are expected to be helpful, honest, and harmless AI assistants~\citep{ouyang2022training,openai2023gpt}. 
Central to these roles, being ``helpful'' is the most fundamental requisite, emphasizing that LLMs should help users to complete various tasks, known as the ``instruction following'' capability. 
Many studies~\citep{wang2022self,zhou2023lima} show that \textit{instruction tuning}, also called supervised fine-tuning, is critical to acquire such capability, by fine-tuning pre-trained models on high-quality prompt-response pairs. 
However, the impact of instruction tuning on the helpfulness of language models remains largely unexplored, limiting the development of better AI assistants.

In this work, we focus on exploring \textit{how instruction tuning changes pre-trained models}. 
Specifically, how do instruction-tuned models utilize the instruction words to guide their generation in a way that differs from pre-trained models?
Step further, how do self-attention heads and feed-forward networks contribute to this difference by adapting their pre-trained knowledge, respectively? 

However, technically answering these questions by interpreting LLMs is non-trivial.
For the first question, we aim to quantify the importance of prompt words to response words, known as attribution explanations. 
Existing work~\citep{selvaraju2016grad,sundararajan2017axiomatic,mu2020compositional} is proposed for the classification problems, which is not suitable for auto-regressive LLMs. 
For the second question, we seek to interpret both self-attention and feed-forward layers within LLMs. 
A straightforward method~\citep{dar2022analyzing,geva2021transformer}, which projects weight vectors into the word embedding space and then selecting the most activated words as explanations, is compromised by the polysemic nature of model weights~\cite{arora2018linear,scherlis2022polysemanticity}, leading to unclear and not concise explanations. 
Other researchers have studied internal activations of the models, such as heatmap visualization~\cite{vig2019bertviz}, sparse auto-encoder decomposition~\cite{brickentowards,cunningham2023sparse}, and knowledge probing~\cite{belinkov2018evaluating,jawahar2019does}, while they may yield biased explanations due to the potential bias in the chosen samples for collecting activations. 
Overall, existing explanation methods cannot be directly applied to auto-regressive LLMs.

To fill these gaps, we first develop a series of explanation methods as a \textit{toolbox} to study LLMs, including a gradient-based method for prompt-response attributions, and techniques to interpret the patterns and concepts in self-attention heads and feed-forward networks at a human-understandable level.
We then investigate the impact of instruction tuning by comparing the explanations coming from the pre-trained and instruction-tuned models. 
This approach provides an internal perspective of exploring instruction tuning, distinguishing it from existing research that primarily focuses on comparing the performance of the model trained under different settings~\cite{liang2023exploring,kung2023models,zhou2023lima,kirk2023understanding}.
We obtain three main findings of the impact of instruction tuning as follows:

\begin{itemize}
\vspace{-0.1cm}
    \item 
    \noindent\textbf{Finding 1:} \textit{It enables models to recognize instruction words in user prompts and drives the generation process to be consistently conditioned on these words.} 
    We introduce a normalization strategy to make the traditional gradient-based methods suitable for attributing response words to prompt words. 
    We observe that instruction words, such as ``Fix grammar errors:'', influence multiple response words across different positions, unlike other words that have a limited effect on the response (Sec.~\ref{sec_token_attribution}). 
    Additionally, we leverage a density function to aggregate the overall importance of each individual prompt word. This importance density score is quantitatively shown to correlate strongly with the models' ability to follow instructions (Sec.~\ref{sec_token_quantitative}).

    \item
    \noindent\textbf{Finding 2:} \textit{It encourages self-attention heads to learn more word relations with instruction verbs than common verbs.}
    We suggest extracting word-word patterns under the local co-occurrence assumption to alleviate the polysemantic challenge in interpreting self-attention heads (Sec.~\ref{sec_att_method}).
    We notice a significant change in the word-word patterns within the same self-attention head after instruction tuning. 
    Analysis shows that the word-word patterns associated with instruction verbs become more popular, especially in the bottom and middle layers, while patterns linked to commonly used verbs do not display a similar increase in popularity.
    This finding demonstrates that self-attention heads have a direct influence on understanding user instructions.

   \item \noindent\textbf{Finding 3:} \textit{It adapts the pre-trained knowledge encoded by feed-forward networks into user-oriented tasks without changing their linguistic structures.}
   We propose interpreting the principal components of weight vectors to reach a ``concept'' level explanation of feed-forward networks (Sec.~\ref{sec_linear_interpret}).
    Our analysis of these concepts spans two dimensions: user-oriented tasks\footnote{User-oriented tasks include ``writing'', ``coding'', ``translation'', and ``solving math problem''.} and linguistic levels\footnote{Linguistic levels include ``phonology'', ``morphology'', ``syntax'', and ``semantic''.}~\citep{thomas2005illuminating}. 
   We find that the proportion of concepts that are suitable for specific tasks, such as writing, coding, and solving math problems, becomes significantly greater after instruction tuning. 
   In contrast, the distribution of these concepts across different linguistic levels remains the same. 
   This phenomenon shows that feed-forward networks adapt their pre-trained knowledge to downstream tasks by slightly rotating the basis of their representation space.

\end{itemize}

This study reveals that instruction words are crucial to instruction-tuned models because of their consistent impact on the generation process, and further emphasizes the distinctive contributions of self-attention mechanisms and feed-forward networks to this functionality.
While our focus is on behavior shifts after instruction tuning, future research might also apply our toolbox to understand LLMs for various other purposes.

\section{Related Work}

\paragraph{Interpreting Language Models.} 
Majority of investigations in interpreting LLMs aimed to understand the decision-making processes of LLMs for a specific task or dataset, which involves feature attribution methods~\citep{li2015visualizing,kokalj2021bert}, attention-based methods~\citep{vig2019bertviz,barkan2021grad}, and sample-based methods~\citep{kim2018interpretability,wu2021polyjuice}.  
Recently, many researchers turned to understanding why LLMs can perform in-context learning~\citep{
xie2021explanation,olsson2022context,li2023towards,wei2023larger,varshney2023stitch,xiong2023can,duan2023shifting}.
In parallel, some works delved into interpreting the internal components of LLMs, including the self-attention mechanism~\citep{elhage2021mathematical,sukhbaatar2019augmenting} and feed-forward networks~\citep{petroni2019language,geva2020transformer,voita2023neurons,huang2023rigorously}.
Our work builds on these foundations, introducing novel interpretation methods tailored for modern LLMs.

\paragraph{Interpreting Instruction-tuned Models.}
Interpreting instruction tuning is still in the early stages of exploring unexpected phenomena. 
A notable example is the ``lost-in-the-middle'' effect identified by \citep{liu2023lost}, which demonstrates that inserting contents in the middle of prompts often results in poor model performance.
Similarly, \cite{zhou2023lima} showed that even only 1000 prompt-response pairs could significantly enhance the instruction-following capabilities of LLMs. 
Moreover, researchers~\citep{liang2023exploring,kung2023models,zhou2023lima} find that instruction-tuned models just learn superficial patterns through instruction tuning.  
These observations motivate us to investigate the internal changes of instruction-tuned models, aiming to reach a comprehensive understanding that recognizes these diverse phenomena under a unified perspective.

\section{Preliminary}
\subsection{Notations}
\vspace{-0.1cm}
Let $\mathcal{V}$ denote a pre-defined vocabulary set, $X$ is an $N$-length prompting text and $Y$ is a $M$-length response from a transformer-based language model $f$, where each token $x_n\in X$ or $y_m\in Y$ comes from $\mathcal{V}$. 
$f$ is defined in a $D$-dimensional space, starting with an input word embedding $\mathbf{E}_i\in\mathbb{R}^{|\mathcal{V}|\times D}$ presenting input tokens in $\mathbf{X}\in\mathbb{R}^{N\times D}$.  
\(\mathbf{X}\) goes through \(L\) transformer blocks, each containing a self-attention module and a feed-forward network. Every self-attention module includes \(H\) heads that operate in a space with \(D'\) dimensions.
Each self-attention head captures word relations by $\mathbf{A}^h=\text{softmax}(\mathbf{X}\mathbf{W}_q^h (\mathbf{X}\mathbf{W}_k^h)^\top / \epsilon)$, where $\mathbf{W}^h_q,\mathbf{W}^h_k\in\mathbb{R}^{D\times D'}$ and $\epsilon$ is a constant. 
The aggregation of heads' outputs is $[\mathbf{A}^1\mathbf{X}\mathbf{W}_v^1;...;\mathbf{A}^H\mathbf{X}\mathbf{W}_v^H]\mathbf{W}_o$.
Each feed-forward network is defined as $\sigma(\mathbf{X}\mathbf{W}_u^\top)\mathbf{W}_p$, where $\sigma$ is a non-linear function, and $\mathbf{W}_u,\mathbf{W}_p\in\mathbb{R}^{D''\times D}$.
Finally, we compute inner products between the processed embeddings and output word embeddings $\mathbf{E}_o\in\mathbb{R}^{ |\mathcal{V}|\times D}$ for next-word prediction.

\subsection{General Experimental Settings}
\vspace{-0.1cm}
\label{sec_general_settings}
\paragraph{Language Models.}
We choose the LLaMA family~\cite{touvron2023LLaMA} as the main subject of this study since LLaMA is one of the most advanced public pre-trained language model families, serving as a foundation of various instruction-tuned models. In addition, we also conduct experiments with the Mistral family~\cite{jiang2023mistral} to demonstrate the generalizability of our findings.  
Specifically, we consider Vicuna~\citep{zheng2023judging} and Mistral-Instruct~\cite{jiang2023mistral} as the instruction-tuned models, while LLaMA~\citep{touvron2023LLaMA} and Mistral~\cite{jiang2023mistral} as their corresponding pre-trained models\footnote{We implement these models with the code and checkpoints available from Huggingface library~\citep{wolf2019huggingface}. We use lmsys/vicuna-7b-delta-v1.1 for Vicuna and mistralai/Mistral-7B-Instruct-v0.1 for Mistral-Instruct.}. 
The greedy search (for reproduction) is employed to generate up to 300 tokens for each input prompt. 

\vspace{-0.1cm}
\paragraph{Instruction Datasets.} We collect user-oriented prompting texts from three publicly available datasets: Self-Instruct~\citep{wang2022self}, LIMA~\citep{zhou2023lima}, and MT-Bench~\citep{zheng2023judging}. 
The Self-Instruct dataset includes 252 pairs of prompts and responses written by humans, used both for generating more pairs and as a test set. LIMA, mainly based on questions and answers from online platforms like Stack Exchange, has 1000 training pairs and 300 testing pairs. On the other hand, MT-Bench, intended only for machine evaluation, has 80 human-written pairs across eight categories but lacks a training set.  
Our analysis focuses on the test sets from these datasets.

\section{Impact of User Prompts for \,\,\,\,\,\,\,\,\,\,\,\, Human Alignment} 
This section focuses on the differential functionalities of user prompts between the instruction-tuned and pre-trained models.
We introduce a gradient-based attribution approach in Sec.~\ref{sec_token_qualitative} to measure the importance of individual input words on specific output words. In Sec.~\ref{sec_token_quantitative}, we compare word importance densities across various models to study their distinctions in using user prompts.

\subsection{Quantifying Prompt Influence on Generation Process}
\label{sec_token_qualitative}
\paragraph{Method.}
\label{sec_token_attribution}
We aim to measure the importance of each prompt token to each response token. In classification, input feature importance is typically measured by monitoring confidence changes upon its removal~\citep{ribeiro2016should,feng2018pathologies}.
Thus, treating text generation as a sequence of word classification tasks, the importance of an input token to an output token is gauged by examining confidence changes in output generation if the input token is removed. 
Therefore, we define \textit{importance} \(I_{n,m}\) of input token \(x_n\) to output token \(y_m\) as:
\begin{equation}
    I_{n,m} = p(y_m|Z_{m}) - p(y_m|Z_{m,/n}), 
    \label{eq_importance}
\end{equation}
where $Z_m$ is the context to generate $y_m$ by concatenating the prompt $X$ and the first $m-1$ tokens of response $Y$, $Z_{m,/n}$ omits token \(x_n\) from \(Z_m\), and $p(\cdot|\cdot)$ is the conditional probability computed by language model $f$. 
We accelerate Eq.~\eqref{eq_importance} with the first-order approximation: 
$I_{n,m} \approx \frac{\partial f(y_m|Z_m)}{\partial\,\mathbf{E}_i[x_n]}\cdot \mathbf{E}_i[x_n]^\top$,
where $\mathbf{E}_i[x_n]$ is the input word embedding of token $x_n$ (check Appendix~\ref{proof} for theoretical justification). 
However, the importance of input tokens cannot be compared across different output tokens due to its dependency on confidence $f(y_m|Z_m)$. It's crucial to recognize that a word with a lower confidence doesn't necessarily imply it is a trivial word. 
Specifically, in language modeling, the likelihood of a word $y$ given previous context $x$ could be extended with Bayes' theorem as $p(y|x)\propto p(x|y)\cdot p(y)$. Here, semantic (non-trivial) words have a lower prior probability $p(y)$ since they are less common in the general corpus. Additionly, models tend to estimate a lower conditional probability $p(x|y)$ since it is more challenging to predict such meaningful words unless they observe a very strong semantic relation. 
Consequently, models are typically more confident about common, less meaningful words, and less confident about semantically rich, rare words.
Thus, we propose to rescale the scores derived from the same output token to ensure their comparability across different output tokens. In addition, we introduce a sparse operation over the rescaled importance to overlook the noise introduced by first-order approximation.
Finally, the normalized pairwise score is
\begin{equation}
S_{n,m} = 
\begin{cases} 
\widetilde{S}_{n,m} & \text{if } \widetilde{S}_{n,m} > b \\
0 & \text{otherwise}
\end{cases},
\end{equation}
where $\widetilde{S}_{n,m}=\left\lceil \frac{L \times I_{n,m}}{\max_{n^\prime=1}^N I_{n^\prime,m}}\right\rceil$, $\lceil\cdot\rceil$ is the ceiling function, and $b\in[0,L]$ is a hyper-parameter determining the minimal interested importance level.

\begin{figure}
  \begin{center}
    \includegraphics[width=1.0\linewidth,height=5.5cm,trim={0.0cm 0.cm 0.0cm 0cm},clip]{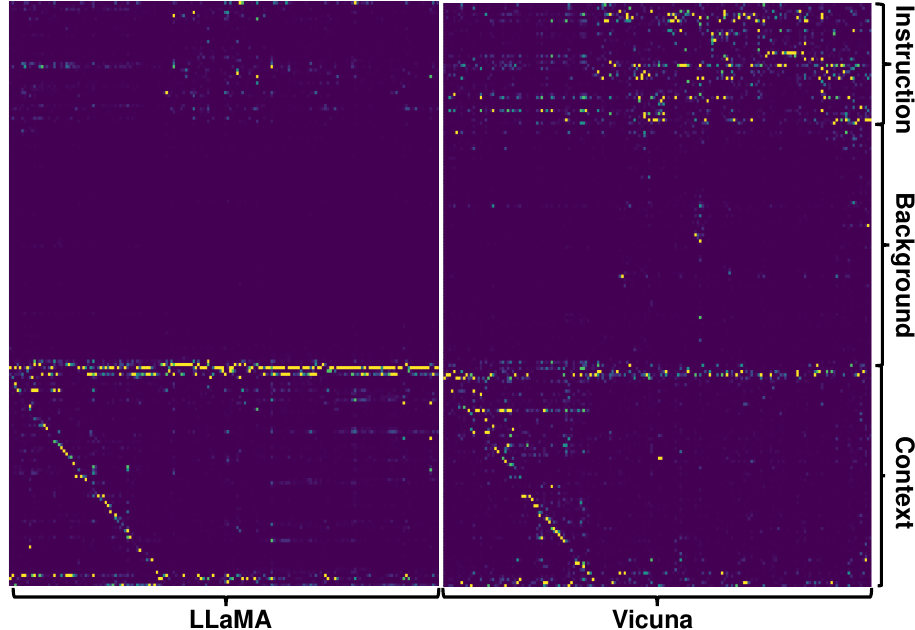}
  \end{center}
  \vspace{-0.5cm}
  \caption[Caption for LOF]{\small Salient maps of the prompt-response pair\protect\footnotemark from LLaMA (left) and Vicuna (right).}
  \vspace{-0.5cm}
  \label{fig_good_cases}
\end{figure}

\paragraph{Settings.}
This qualitative experiment demonstrates how prompt words contribute to response generation via visualizing salient maps based on normalized pairwise importance $S_{n,m}$. 
We set $L=10$ and $b=0$ for visualization. 
Figure~\ref{fig_good_cases} provides a pair of salient maps to the same prompt corresponding to the model-generated responses from LLaMA and Vicuna, respectively. We show more visualization cases in Appendix~\ref{appendix_salientmaps}. 

\label{sec_impacts_of_instructions}
\noindent

\vspace{-0.2cm}
\noindent\textbf{Obs-1: Instruction tuning helps the models distinguish between instruction and context words more accurately.}
We provide a visualization case\footnotetext{\tiny The prompt boldfaces its direct \textbf{instruction words} and underlines its \underline{background}: \tiny \textbf{Analyze the word choice, phrasing, punctuation, and capitalization in the given email. How may the writer of this email sound to the reader?} \underline{These tones include Disheartening, Accusatory, Worried, Curious, Surprised, Disapproving,} \underline{\noindent Unassuming, Formal, Assertive, Confident, Appreciative, Concerned, Sad, Informal, Regret} \underline{\noindent ful, Encouraging, Egocentric,} \underline{ Joyful, Optimistic, and Excited.}\textbackslash n\textbackslash nInput: Hi Jen, \textbackslash nI hope you're well. Can we catch up today? I'd appreciate your input on my presentation for tomorrow's meeting. I'd especially love it if you could double-check the sales numbers with me. There's a coffee in it for you!\textbackslash n\textbackslash nOutput: 
} that asks the models to analyze the tone (instruction) of a given email (context) into one of the listed categories (background). 
Both models begin their responses by repeating the email. Later, Vicuna successfully analyzes the tone of the email, while LLaMA fails to do that. 
Figure~\ref{fig_good_cases} (right) shows that the instruction part is generally brighter than the background and context parts, indicating a strong influence of instruction words in shaping response generation. In contrast, context lines only light up in specific spans and show a diagonal pattern at the left button of both figures (models are repeating the email). 
The differences between the left and right plots highlight the impact of instruction tuning. 
Specifically, the left plot has certain context lines that appear less bright in the right plot, while certain instruction lines in the right plot stand out more. 
This visualization raises a hypothesis that the instruction words \textit{constantly} contribute to the response generation if the model successfully follows the user intention. Sec.~\ref{sec_token_quantitative} will quantitatively verify this hypothesis.

\subsection{Assessing Instruction Following Capability with Importance Density} 
\vspace{-0.05cm}
\label{sec_token_quantitative}

\begin{table}
\small
\centering
  \caption{Importance density on instruction words over followed and unfollowed instances from Vicuna.} 
  \small
  \vspace{-0.3cm}
  \label{fig_distribution_of_giascores}
  \begin{tabular}{cccc}
    \hline
    \hline
    \textbf{Dataset} & \textbf{Followed} & \textbf{Unfollowed} & \textbf{p-value} \\
    \hline
    Self-Instruct & $1.2283_{\pm0.52}$ & $0.8917_{\pm0.48}$ & $1.4e^{-4}$ \\
    LIMA & $1.6173_{\pm0.47}$ & $1.2799_{\pm0.44}$ & $4.3e^{-6}$ \\
    MT-Bench & $1.4584_{\pm0.55}$ & $0.9290_{\pm0.53}$ & $2.3e^{-4}$ \\
    \hline
    \hline
\end{tabular}
\vspace{-0.5cm}
\label{table_instruction_following}
\end{table}

\paragraph{Method.}
\label{sec_overall_attribution}
We aim to measure the overall attribution of each input token to the entire response generation process. 
Based on Sec.~\ref{sec_impacts_of_instructions}, an input token should acquire a greater attribution score if it is important to generate more output tokens.
Following this intuition, the input token $x_n$'s attribution $a_n$ is measured by leveraging $\boldsymbol{\ell}_1/\boldsymbol{\ell}_p$ density function over the normalized importance to all output tokens: $a_n = ||S_{n}||_1 / ||S_{n}||_p$, 
where $S_n=[S_{n,1},...,S_{n,M}]$, and $p\in \mathbb{R}^+$ serves as a hyper parameter. 
One nice property of this density function is if two input tokens have the same total importance, then the one having greater maximum importance would receive a greater density score (check \citep{hurley2009comparing} for proof). 

\paragraph{Settings.} 
This experiment quantitatively justifies the assumption observed from Sec.~\ref{sec_impacts_of_instructions} that a model aligns with human intention if it constantly uses instruction words to guide the generation.
Specifically, we manually annotate a dataset, where each prompt has been marked its instruction part, and each response is labeled as either ``followed'' or ``unfollowed''. 
Please check Appendix~\ref{appendix_human_annotations} for the annotation details.
Here, the instruction part includes sentences that describe background information and actions for a task. 
On the other hand, ``followed'' indicates that the model provides information pertinent to the user intention, regardless of the response's factual correctness. 
For each prompt-response pair sourced from our datasets, we compute the importance density score with $L=10$, $b=7$, and $p=4$. 
We further normalize the scores to ensure comparability across different instances and remove the instances with a short response (less than 5 tokens) as their estimations of density are not stable. 
Table~\ref{table_instruction_following} compares the average importance densities between the followed and unfollowed instances from Vicuna, while Table~\ref{tbl_distribution_of_giascores_vicuna_llama} compares the average importance densities between the Vicuna generated or LLaMA generated instances.  


\vspace{0.2cm}
\noindent\textbf{Obs-2: The importance density on instruction words reflects the models' behaviors in following user intentions.}
From Table~\ref{table_instruction_following}, it becomes evident that attribution scores for ``followed'' instances consistently outperform those of ``unfollowed'' across all datasets. This distinction is statistically validated by notably low p-values, where the null-hypothesis is the average importance densities of followed instances are greater than that of unfollowed, underscoreing a strong correlation between the importance density scores on instruction words and the instruction following capability.
Case studies in Appendix~\ref{appendix_false_negative} suggest that instruction-tuned models may pretend to follow instructions without realizing user instructions. 
Appendix~\ref{sec_token_macroscopic} further emphasizes that this observation cannot be simply explained by the fact that instruction words usually appear at the heads or tails of prompts.

\begin{table}
\small
\centering
  \caption{Importance density on instruction words over responses generated by Vicuna and LLaMA.}
  \small
  \vspace{-0.3cm}
  \label{tbl_distribution_of_giascores_vicuna_llama}
  \begin{tabular}{cccc}
    \hline
    \hline
    \textbf{Dataset} & \textbf{Vicuna} & \textbf{LLaMA}& \textbf{p-value} \\
    \hline
    Self-Instruct & $1.1302_{\pm0.54}$ & $0.9394_{\pm0.48}$ & $1.9e^{-5}$ \\
    LIMA & $1.5579_{\pm0.49}$ & $1.2683_{\pm0.43}$ & $\,\,2.5e^{-14}$ \\
    MT-Bench & $1.3440_{\pm0.60}$ & $1.1777_{\pm0.58}$ & $0.0382$ \\
    \hline
    \hline
\end{tabular}
\vspace{-0.5cm}
\label{table_instruction_following_vicuna_llama}
\end{table}

\vspace{0.2cm}
\noindent\textbf{Obs-3: Instruction-tuned models archive a greater importance density than their pre-trained counterparts.}  
Table~\ref{table_instruction_following_vicuna_llama} reports the average importance density over the instruction words by giving responses generated by Vicuna or LLaMA. 
We observe that Vicuna constantly assigns denser importance scores on the instruction words compared to LLaMA across the three datasets, where two out of them are statistically significant ($p<0.05$). Here, the null hypothesis is that the average importance densities computed by responses generated by Vicuna are greater than that of LLaMA. 
According to Obs-2, we draw a conclusion that Vicuna demonstrates a better instruction-following capability than LLaMA by more accurately identifying instruction words and then successfully using them to guide its response generation process.

\section{Shift within Instruction-tuned Models}
\vspace{-0.3cm}
\label{sec_evolution}
This section studies the distinctive contributions of components in LLMs for human alignment. The self-attention heads and feed-forward networks are discussed in Sec.~\ref{sec_att_method} and~\ref{sec_linear_interpret}, respectively.

\subsection{Analyzing Self-Attention Heads}
\label{sec_att_method}
\paragraph{Method.}
We aim to interpret the behavior of self-attention heads with word pairs. 
Let $\mathbf{W}[d]$ denote the $d$-th row of matrix $\mathbf{W}$. 
Given a self-attention head with weights $\mathbf{W}^h_q,\mathbf{W}^h_k$, the relation between a pair of words $(w_a, w_b)$ is computed as $\mathbf{A}_{a,b} \propto \sum_{d=1}^D \mathbf{E}_i[w_a](\mathbf{W}^{h\top}_q[d])^\top \times \mathbf{E}_i[w_b](\mathbf{W}^{h\top}_k[d])^\top$.
~\citet{dar2022analyzing} computes the attention scores between words from the entire vocabulary $\mathcal{V}$ based on $\mathbf{W}^h_q$ and $\mathbf{W}^h_k$, and selects the top-$K$ word pair with the greatest scores to interpret the $h$-th self-attention head. 
However, we notice that the word pairs obtained by this approach are redundant, leading to a less comprehensive understanding of the self-attention head. 
To overcome this problem, we propose to interpret a self-attention head by aggregating the word pairs activated by its neuron pairs, which is motivated by the fact that the relation $\mathbf{A}_{a,b}$ linearly relates to the activations of column vectors of weights $\mathbf{W}_q^h$ and $\mathbf{W}_k^h$, called ``neurons'' in this paper. 
But this approach suffers from the polysemantic nature of neurons~\cite{elhage2022toy,anthropic2023towards}, introducing word pairs that are meaningless to be connected. 
Considering the self-attention mechanism is designed for capturing word relations within the same input texts, we introduce a co-occurrence constraint to form the word pairs. 
Specifically, we first interpret neurons $\mathbf{W}^{h\top}_q[d]$ and $\mathbf{W}^{h\top}_k[d]$ by collecting the top-$K$ words that could most activate them, i.e., $\mathcal{E}_q^d = \arg max_{\mathcal{V}'\subseteq\mathcal{V}, |\mathcal{V}'|=K} \sum_{w\in\mathcal{V}'}\mathbf{E}_i[w]\cdot (\mathbf{W}^{h\top}_q[d])^\top$ and $\mathcal{E}_k^d = \arg max_{\mathcal{V}'\subseteq\mathcal{V}, |\mathcal{V}'|=K} \sum_{w\in\mathcal{V}'}\mathbf{E}_i[w]\cdot (\mathbf{W}^{h\top}_k[d])^\top$.   
We then form the word pair list $\mathcal{E}_{qk}^d = \{(w_q, w_k): cos(\mathbf{e}_q, \mathbf{e}_k) > \theta\}$ for $w_q\in\mathcal{E}_q^d$, $w_k\in\mathcal{E}_k^d$,
where $\mathbf{e}_q$, $\mathbf{e}_k$ are their GloVe embeddings~\citep{pennington2014glove}, and $\theta$ is a threshold. 
The final explanation of a self-attention head is derived from its frequent neuron-level word pairs. 

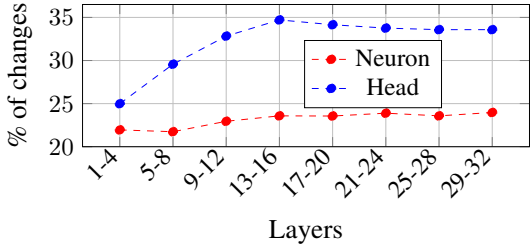
\begin{figure}
  \begin{center}
    \begin{tikzpicture}[scale=0.85]
    \begin{axis}[
        xtick={1,2,3,4,5,6,7,8},
        xticklabels={1-4,5-8,9-12,13-16,17-20,21-24,25-28,29-32},
        xticklabel style={rotate=45, anchor=east}, 
        xlabel={Layers},
        xlabel style={font=\large},
        ylabel style={font=\large},
        ylabel={\% of changes},
        ymin=20, ymax=36.5,
        height=3.8cm,
        width=8.5cm,
        range={(0.1, 1.0},
    legend style={
        at={(0.65,0.75)},   
        anchor=north      
    },
        grid=major,
    ]

    \addplot[color=red,dashed,mark=*] coordinates {(1,21.95) (2,21.74) (3,22.95) (4,23.58) (5,23.56) (6,23.89) (7,23.58) (8,23.97)};
    \addlegendentry{Neuron}

    \addplot[color=blue,dashed,mark=*] coordinates {(1,24.99) (2,29.57) (3,32.83) (4,34.71) (5,34.15) (6,33.76) (7,33.58) (8,33.58)};
    \addlegendentry{Head}

    \end{axis}
\end{tikzpicture}
  \end{center}
  \vspace{-0.6cm}
  \caption{Differences of word-word patterns between Vicuna and LLaMA over neuron and head levels.}
  \vspace{-0.4cm}
  \label{fig_bilinear_knowledge_dist}
\end{figure}

\paragraph{Settings.}
We consider $K=100$ as a constant and $\theta$ as dynamic values for different words. 
Specifically, we first compute the cosine similarity between the given word and 1000 frequent words with their GloVe word embeddings. 
The threshold of a word is its average similarity adding 1.96 times its standard deviation, and the greater one of two words is that of a word pair. 
We conduct a qualitative analysis of the word pairs in Appendix~\ref{appendix_att}. 

The impact of instruction tuning on self-attention heads is studied by comparing the word pair lists from the pre-trained and tuned models. First, we quantify the changes of word pair lists with the intersection rate $M=\frac{\mathcal{E}_{pt}\cap \mathcal{E}_{ft}}{\mathcal{E}_{pt}\cup \mathcal{E}_{ft}}$, where $\mathcal{E}_{pt}$ and $\mathcal{E}_{ft}$ denote the top-100 word pairs of the pre-trained and tuned models. 
Figure~\ref{fig_bilinear_knowledge_dist} visualizes $1-M$ over various layer groups.
We also investigate how these changed word pairs related to the instruction-following capability, focusing on verbs. 
Specifically, We identify 45 instruction verbs (e.g., ``write'', ``create'', and ``classify'') based on~\cite{wang2022self,ouyang2022training}, and also assemble a control set of 3000 frequent verbs~\cite{robyn_speer_2022_7199437}.
For each verb, we calculate the proportion of the self-attention head that encodes more word pairs for that verb after instruction tuning.
We only consider those self-attention heads that change in the number of word pairs for the verb after instruction tuning and report the results on Table~\ref{tbl_att_instruct_word_freq}.

\vspace{0.2cm}
\noindent\textbf{Obs-4: Instruction tuning significantly modifies self-attention heads.}
Figure~\ref{fig_bilinear_knowledge_dist} shows that as layer depth increases, the differences between word pair lists become more significant. 
This not only illustrates the significant impact of instruction tuning on the self-attention heads, but also shows that the proposed method can capture the diverse word-word relationships encoded by the self-attention layer.

\begin{table}
\small
\centering
  \caption{Percentage of self-attention heads encoding certain verbs after instruction tuning.} 
  \vspace{-0.2cm}
  \small
  \label{tbl_att_instruct_word_freq}
  \begin{tabular}{@{}c|cccc@{}}
    \hline
    \hline
    \textbf{Family} & \textbf{Layers} & \textbf{Instruct} & \textbf{General} & \textbf{p-value} \\
    \hline
    \multirow{4}{*}{LLaMA} & 1-8 & $65.96_{\pm 26.96}$ & $49.61_{\pm 31.89}$ & $0.0050$\\
                           & 9-16  & $62.30_{\pm 32.80}$ & $50.72_{\pm 32.26}$ & $0.1330$\\
                           & 17-24  & $52.15_{\pm 39.44}$ & $50.87_{\pm 30.35}$ & $0.8785$\\
                           & 25-32  & $43.49_{\pm 35.82}$ & $48.71_{\pm 29.96}$ & $0.4856$\\
    \hline
    \multirow{4}{*}{Mistral} & 1-8 & $68.92_{\pm 30.18}$ & $52.72_{\pm 28.59}$ & $0.0114$\\
                           & 9-16  & $55.62_{\pm 41.82}$ & $51.84_{\pm 27.24}$ & $0.6700$\\
                           & 17-24  & $66.92_{\pm 31.75}$ & $52.95_{\pm 28.44}$ & $0.0643$\\
                           & 25-32  & $57.04_{\pm 38.75}$ & $52.13_{\pm 27.79}$ & $0.5697$\\
    \hline
    \hline
\end{tabular}
\vspace{-0.5cm}
\end{table}

\vspace{0.2cm}
\noindent\textbf{Obs-5: Instruction tuning encodes more instruction verbs in lower self-attention heads.}
Table~\ref{tbl_att_instruct_word_freq} demonstrates that instruction tuning notably increases the propensity of self-attention heads of LLaMA family, particularly in lower (1-8) and middle (9-16) layers, to encode word-word patterns associated with instruction verbs. This enhancement is statistically significant ($p<0.05$) within the first 8 layers. In contrast, approximately 50\% of self-attention heads exhibit a similar tendency for general verbs, while 50\% refers to a neutral impact, signifying neither an increase nor decrease in word relations for the given verbs. 
On the other hand, all layers of the Mistral family encoded more word pairs with instruction verbs, which may be the reason for the stronger instruction-following ability of Mistral.
Overall, this difference indicates that instruction tuning teaches self-attention to identify various detailed instructions.

\subsection{Analyzing Feed-forward Networks}
\label{sec_linear_interpret} 
\paragraph{Method.} 
We aim to interpret the knowledge of feed-forward networks in the \textit{concept} level. 
We treat each feed-forward network $\sigma(\mathbf{X}\mathbf{W}_u^\top)\mathbf{W}_p$ as key-value memories~\citep{geva2020transformer}, where each row vector of $\mathbf{W}_u$ and $\mathbf{W}_p$ stores a textual pattern. 
However, these textual patterns (neurons) are usually polysemantic~\citep{elhage2022toy,anthropic2023towards}, causing each textual pattern not to be interpreted within a concise meaning~\cite{geva2021transformer}.
Thus, we propose to seek a set of orthogonal vectors that capture the major directions in which these patterns spread. 
Formally, given patterns $\mathbf{W}_p$, we construct the covariance matrix as $\mathbf{C}=\widetilde{\mathbf{W}}_p^\top\widetilde{\mathbf{W}}_p$, where $\widetilde{\mathbf{W}}_p$ is the centralized matrix of $\mathbf{W}_p$ with zero-mean columns. 
Then the orthogonal basis vectors $\mathbf{V}$ of these patterns satisfy:
\begin{equation}
\label{eqn:pca_basis}
\mathbf{C}\mathbf{V} = \mathbf{\Lambda}\mathbf{V},
\end{equation}
where each column vector of $\mathbf{V}\in\mathbb{R}^{D\times D}$ has unit length, $\mathbf{\Lambda}=diag([\lambda_1,...,\lambda_D])$, and $\lambda_1\geq...\geq\lambda_D\geq0$.  
In this context, our primary focus lies on the top-$R$ values of $\mathbf{\Lambda}$ along with their corresponding column vectors in $\mathbf{V}$. 
This is due to the fact that they show the \textit{principal directions} of the encoded patterns from $\mathbf{W}_p$. 
We then project word embeddings $\mathbf{E}_o$ to each principal directions and find the top-$K$ relevant words for interpretation: $\mathcal{E}_r = \arg max_{\mathcal{V}'\in\mathcal{V}, |\mathcal{V}'|=K} \sum_{w\in\mathcal{V}'}\mathbf{V}^\top[r]\mathbf{E}_o[w]$, 
where $\mathbf{V}^\top[r]$ is the $r$-th column vector of $\mathbf{V}$, $\mathbf{E}_o[w]$ is the output word embedding of $w$. 
Since $\mathbf{V}^\top[r]$ is a unit vector, $\mathbf{V}^\top[r]\mathbf{E}_o[w]$ measures the projection length of the word vector in this direction. 
Thus, it is natural to represent this vector with the words having the largest projection length, and the word list could be further summarized as a textual description by a human or a machine annotator. 

\paragraph{Settings.}
\label{sec_second_category}
We create a new vocabulary derived from ShareGPT~\citep{sharegpt} to make the candidate words $\mathcal{V}$ more understandable compared to a large number of sub-tokens from the built-in LLaMA vocabulary.
We then analyze the first 300 basis vectors of each feed-forward network from LLaMA and Vicuna with their top 15 relevant words. 
ChatGPT~\footnote{We employ ChatGPT-turbo-3.5-0613 in this work.} is considered our machine annotator for this experiment.
Table~\ref{tbl_ffn_wordlists} provides sample word lists and their descriptions. More cases are available in Appendix~\ref{appendix_ffn}. The detailed settings and statistics of concept descriptions are shown in Appendix~\ref{appendix_summarization_results}. 
We discuss the results of the principal components in Appendix~\ref{appendix_pca}.

To study the evolution of pre-trained knowledge, we condense tasks from previous research~\citep{zheng2023judging,ouyang2022training} to scenarios including writing, math, coding, and translation. We then identify which scenarios a concept could be used for (see Appendix~\ref{appendix_chatgpt}). 
Note that some concepts may fit multiple scenarios. Also, we sort concepts into phonology~\footnote{Phonology studies sound systems, e.g. words with \textit{``le'' sound}: brittle, tackle, chuckle, pickle.}, morphology~\footnote{Morphology studies word structure, e.g. words with \textit{``sub-'' prefix}: subarray, subculture, subway.}, syntax, or semantics linguistic levels based on the disciplines in the linguistic subject~\citep{thomas2005illuminating}. Table~\ref{table_task_concept} shows the percentage of knowledge for different scenarios and linguistic levels.

\begin{table}
\small
\centering
  \caption{Interpreting the last feed-forward network of Vicuna with the proposed decomposition method.} 
  \vspace{-0.2cm}
  \small
  \label{tbl_ffn_wordlists}
  \begin{tabular}{l|l}
    \hline
    \hline
    \textbf{Description} & \textbf{Words} \\
    \hline
     medical abbreviation &  \small{CBT, RTK, RT, RH, HRV, MT, ...}  \\ 
      starting with ``the''  & \small{the, theological, theology, ...}     \\
     hyphenated terms & \small{one-of-a-kind, state-of-the-art, ...} \\
      numbers & \small{sha256, tt, 8266, 768, 1986, ...}       \\  
    \hline
    \hline
\end{tabular}
\vspace{-0.5cm}
\end{table}

\vspace{0.5cm}
\noindent\textbf{Obs-6: The principal vectors of the weights of feed-forward networks provide concept-level understandings of the encoded knowledge.}
We select four representative principal components and their explanations from the last feed-forward network of Vicuna and display them in Table~\ref{tbl_ffn_wordlists}.
More cases are available in Tables~\ref{tbl:appendix_full_table_ffn_1} and~\ref{tbl:appendix_full_table_ffn_2}.  
In Table~\ref{tbl_ffn_wordlists}, the descriptions of the four principal vectors span diverse topics, ranging from medical (``medical abbreviation'') to linguistic (``starting with {\it the}''). Notably, the concept of medical abbreviations stands out, as it's often difficult for human annotators to discern their medical relevance. This indicates the advantage of utilizing machine annotators for their vast knowledge. 
Coincidentally, Appendix~\ref{appendix_summarization_results} shows that around 60\% of the first 300 principal components from the middle layers of Vicuna could be interpreted by ChatGPT. 
This evidence empirically verifies the rationale for analyzing feed-forward networks with the proposed method.

\begin{table}
\small
\centering
  \caption{Concept distribution of LLaMA family over different user scenarios and linguistic levels.} 
  \vspace{-0.3cm}
  \small
  \label{tbl_distribution_of_scenarios}
  \begin{tabular}{c|cp{1cm}p{1cm}p{1cm}}
    \hline
    \hline
    & \textbf{Category} & \textbf{Vicuna} & \textbf{LLaMA} & \textbf{p-value} \\
    \hline
    \multirow{5}{*}{Scenarios} & Writing & $53.50_{\pm.46}$  & $51.47_{\pm.92}$ & $0.0154$ \\
    & Coding &  $29.45_{\pm.43}$ & $28.64_{\pm.48}$ & $0.0350$\\
    & Math & $5.21_{\pm.36}$ & $5.04_{\pm.33}$ & $0.5193$\\
    & Translation & $25.30_{\pm.39}$ & $26.27_{\pm.70}$ & $0.0411$ \\
    \hline
    \multirow{5}{*}{Linguistic} &  Phonology & $1.18_{\pm.11}$  & $1.15_{\pm.07}$ & $0.6251$ \\
    & Morphology & $17.16_{\pm.49}$  & $16.83_{\pm.60}$ & $0.4223$ \\
    & Syntax & $7.16_{\pm.31}$  & $7.52_{\pm.50}$ & $0.2551$ \\
    & Semantic & $74.70_{\pm.65}$ & $74.66_{\pm.67}$ & $0.9394$ \\
    \hline
    \hline
\end{tabular}
\label{table_task_concept}
\vspace{-0.5cm}
\end{table}

\vspace{0.2cm}
\noindent\textbf{Obs-7: Instruction tuning shifts the principal vectors of feed-forward networks toward user-oriented tasks without moving them across linguistic levels.}
\label{obs_7}
Table~\ref{table_task_concept} reveals that Vicuna encodes more concepts than LLaMA for writing, coding, and math tasks, with the difference in writing and coding being statistic significant ($p < 0.05$), where the null-hypothesis is knowledge proportions of a certain category are equal after instruction tuning. However, the concepts for translation are less after tuning, indicating that multi-linguistic knowledge is forgotten. 
Although we could observe the changes over the user view, it remains the same from the linguistic view. 
In particular, Vicuna and LLaMA show nearly identical distributions across the four linguistic levels without statistically significant ($p>0.05$), suggesting that instruction tuning does not alter the distribution of pre-trained knowledge across linguistic levels. Appendix~\ref{appendix_mistral_ffn} shows the same observation on the Mistral family.

\begin{figure}
    \centering
    \includegraphics[width=0.83\linewidth,trim={0.4cm 0.3cm 0.4cm 0.3cm},clip]{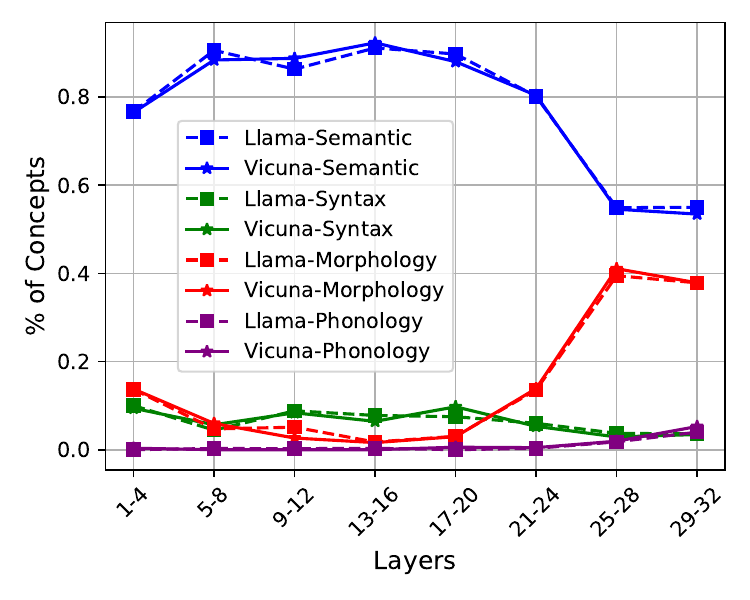}
    \vspace{-0.4cm}
    \caption{Distribution of concepts at linguistic levels over different model layers.}
    \label{fig_ling_knowledge_dist}
    \vspace{-0.4cm}
\end{figure}

\vspace{0.2cm}
\noindent\textbf{Obs-8: The proportion of semantic knowledge first increases and then decreases from bottom to top layers, while that of morphology knowledge does the opposite.}
Figure~\ref{fig_ling_knowledge_dist} displays how concepts from various linguistic levels are spread across layers. First, there isn't a noticeable distribution shift between Vicuna and LLaMA, which matches Obs-7. 
One noteworthy observation is the opposite ``U''-shape trend of semantic knowledge, mirrored by a regular ``U''-shape of morphology. This pattern is surprising, especially since previous studies in computer vision suggest that 
basic features are extracted in the bottom layers, and 
compositional knowledge is learned in the top layers~\citep{zeiler2014visualizing,selvaraju2016grad}. 
However, since LLaMA is a generative model, this unusual pattern makes some sense.
Specifically, we conjecture that LLaMA learns more morphology knowledge (e.g., prefix and suffix patterns) in the last few layers to simulate a prefix-tree structure~\citep{fredkin1960trie,giancarlo1995generalization,paladhi2008generation,shan2012fast}. 
By doing so, LLaMA could use fewer parameters to memorize more phrases to complete the next-word prediction task. 
We leave explorations as future work. 

\section{Discussion}
\vspace{-0.1cm}
Our findings provide a unique perspective to align with recent studies. 1) The importance of \textit{prompt diversity} is highlighted by both us and \cite{zhou2023lima,wang2022self}. Since our three findings suggest that instruction tuning links the pre-trained model to user tasks, we could expect a better alignment with human intentions if the model is exposed to broader prompts. 2) The efficacy of \textit{training self-attention with first priority} (LoRA fine-tuning)~\citep{taori2023alpaca,alpacalora} is corroborated by Finding-1 and Finding-2. Specifically, Finding-1 illustrates the capability to distinguish instruction words is essential to the instruction following, while Finding-2 highlights that self-attention heads directly learn instruction relations. 3) The advantage of \textit{training feed-forward networks} (fully fine-tuning)~\citep{sun2023comparative} is evident from Finding-2 and Finding-3, which demonstrate that feed-forward networks update their knowledge toward user tasks.

\section{Conclusion} 
\vspace{-0.1cm}
This paper presents a comprehensive analysis of instruction tuning for user intention alignment by quantitatively and qualitatively comparing the post-hoc explanations between pre-trained and fine-tuned models. 
Our findings indicate that instruction tuning links the pre-trained model to user intentions, including encoding more instruction words' knowledge within self-attention, and rotating general knowledge from feed-forward networks towards user usage. 
It is worth mentioning that the interpretability toolbox used in this study can also support future general research on LLMs. 

\clearpage
\section{Limitations}
This study aims to investigate the impact of instruction tuning on pre-trained language models in terms of human alignment. 
A primary constraint of this work is that the introduced explanation toolbox is developed on the availability of model weights and gradients, indicating a white-box approach. Consequently, these tools may not be fully effective for analyzing black-box instruction-tuned models, like ChatGPT~\cite{bai2022training} and Claude~\cite{anthropic2023claude}. We will seek to enhance our toolbox by incorporating methods suitable for black-box model analysis in the future. 
On the other hand, another key technology related to human alignment for LLMs is Reinforcement Learning with Human Feedback (RLHF)~\cite{stiennon2020learning,bai2022training}, which is another aspect not touched on in this article. We encourage researchers to apply our toolbox to study RLHF-tuned models and explore the different roles of instruction tuning and RLHF for human alignment.

\section{Ethical Impact}
This research employs the pre-trained models LLaMA~\cite{touvron2023LLaMA} and Mistral~\cite{jiang2023mistral} as well as their variants Vicuna~\cite{zheng2023judging} and Mistral-Instruct~\cite{jiang2023mistral}, under their respective academic-use licenses. The utilization of these models adheres to their specific terms, focusing exclusively on scholarly purposes. Additionally, our study incorporates four datasets: Self-Instruct~\citep{wang2022self}, LIMA~\citep{zhou2023lima}, MT-Bench~\citep{zheng2023judging}, and ShareGPT~\cite{sharegpt}, each under its own usage conditions. These conditions include compliance with privacy and data protection standards. In presenting our findings, we have rigorously ensured that no personal identifiers are disclosed and that the content remains free from offensive material, aligning with ethical research practices.

\bibliography{anthology,custom}

\begin{thebibliography}{66}
\expandafter\ifx\csname natexlab\endcsname\relax\def\natexlab#1{#1}\fi

\bibitem[{Anthropic(2023)}]{anthropic2023claude}
Anthropic. 2023.
\newblock \emph{Model Card and Evaluations for Claude Models}.

\bibitem[{Arora et~al.(2018)Arora, Li, Liang, Ma, and Risteski}]{arora2018linear}
Sanjeev Arora, Yuanzhi Li, Yingyu Liang, Tengyu Ma, and Andrej Risteski. 2018.
\newblock Linear algebraic structure of word senses, with applications to polysemy.
\newblock \emph{Transactions of the Association for Computational Linguistics}, 6:483--495.

\bibitem[{Bai et~al.(2022)Bai, Jones, Ndousse, Askell, Chen, DasSarma, Drain, Fort, Ganguli, Henighan et~al.}]{bai2022training}
Yuntao Bai, Andy Jones, Kamal Ndousse, Amanda Askell, Anna Chen, Nova DasSarma, Dawn Drain, Stanislav Fort, Deep Ganguli, Tom Henighan, et~al. 2022.
\newblock Training a helpful and harmless assistant with reinforcement learning from human feedback.
\newblock \emph{arXiv preprint arXiv:2204.05862}.

\bibitem[{Barkan et~al.(2021)Barkan, Hauon, Caciularu, Katz, Malkiel, Armstrong, and Koenigstein}]{barkan2021grad}
Oren Barkan, Edan Hauon, Avi Caciularu, Ori Katz, Itzik Malkiel, Omri Armstrong, and Noam Koenigstein. 2021.
\newblock Grad-sam: Explaining transformers via gradient self-attention maps.
\newblock In \emph{Proceedings of the 30th ACM International Conference on Information \& Knowledge Management}, pages 2882--2887.

\bibitem[{Belinkov et~al.(2018)Belinkov, M{\`a}rquez, Sajjad, Durrani, Dalvi, and Glass}]{belinkov2018evaluating}
Yonatan Belinkov, Llu{\'\i}s M{\`a}rquez, Hassan Sajjad, Nadir Durrani, Fahim Dalvi, and James Glass. 2018.
\newblock Evaluating layers of representation in neural machine translation on part-of-speech and semantic tagging tasks.
\newblock \emph{arXiv preprint arXiv:1801.07772}.

\bibitem[{Bricken et~al.(2023{\natexlab{a}})Bricken, Templeton, Batson, Chen, Jermyn, Conerly, Turner, Anil, Denison, Askell, Lasenby, Wu, Kravec, Schiefer, Maxwell, Joseph, Tamkin, Nguyen, McLean, Burke, Hume, Carter, Henighan, and Olah}]{anthropic2023towards}
Trenton Bricken, Adly Templeton, Joshua Batson, Brian Chen, Adam Jermyn, Tom Conerly, Nicholas~L Turner, Cem Anil, Carson Denison, Amanda Askell, Robert Lasenby, Yifan Wu, Shauna Kravec, Nicholas Schiefer, Tim Maxwell, Nicholas Joseph, Alex Tamkin, Karina Nguyen, Brayden McLean, Josiah~E Burke, Tristan Hume, Shan Carter, Tom Henighan, and Chris Olah. 2023{\natexlab{a}}.
\newblock \emph{Decomposing Language Models With Dictionary Learning}.

\bibitem[{Bricken et~al.(2023{\natexlab{b}})Bricken, Templeton, Batson, Chen, Jermyn, Conerly, Turner, Anil, Denison, Askell et~al.}]{brickentowards}
Trenton Bricken, Adly Templeton, Joshua Batson, Brian Chen, Adam Jermyn, Tom Conerly, Nick Turner, Cem Anil, Carson Denison, Amanda Askell, et~al. 2023{\natexlab{b}}.
\newblock Towards monosemanticity: Decomposing language models with dictionary learning. transformer circuits thread, 2023.

\bibitem[{Cunningham et~al.(2023)Cunningham, Ewart, Riggs, Huben, and Sharkey}]{cunningham2023sparse}
Hoagy Cunningham, Aidan Ewart, Logan Riggs, Robert Huben, and Lee Sharkey. 2023.
\newblock Sparse autoencoders find highly interpretable features in language models.
\newblock \emph{arXiv preprint arXiv:2309.08600}.

\bibitem[{Dar et~al.(2022)Dar, Geva, Gupta, and Berant}]{dar2022analyzing}
Guy Dar, Mor Geva, Ankit Gupta, and Jonathan Berant. 2022.
\newblock Analyzing transformers in embedding space.
\newblock \emph{arXiv preprint arXiv:2209.02535}.

\bibitem[{Duan et~al.(2023)Duan, Cheng, Wang, Wang, Zavalny, Xu, Kailkhura, and Xu}]{duan2023shifting}
Jinhao Duan, Hao Cheng, Shiqi Wang, Chenan Wang, Alex Zavalny, Renjing Xu, Bhavya Kailkhura, and Kaidi Xu. 2023.
\newblock Shifting attention to relevance: Towards the uncertainty estimation of large language models.
\newblock \emph{arXiv preprint arXiv:2307.01379}.

\bibitem[{Elhage et~al.(2022)Elhage, Hume, Olsson, Schiefer, Henighan, Kravec, Hatfield-Dodds, Lasenby, Drain, Chen et~al.}]{elhage2022toy}
Nelson Elhage, Tristan Hume, Catherine Olsson, Nicholas Schiefer, Tom Henighan, Shauna Kravec, Zac Hatfield-Dodds, Robert Lasenby, Dawn Drain, Carol Chen, et~al. 2022.
\newblock Toy models of superposition.
\newblock \emph{arXiv preprint arXiv:2209.10652}.

\bibitem[{Elhage et~al.(2021)Elhage, Nanda, Olsson, Henighan, Joseph, Mann, Askell, Bai, Chen, Conerly et~al.}]{elhage2021mathematical}
Nelson Elhage, Neel Nanda, Catherine Olsson, Tom Henighan, Nicholas Joseph, Ben Mann, Amanda Askell, Yuntao Bai, Anna Chen, Tom Conerly, et~al. 2021.
\newblock A mathematical framework for transformer circuits.
\newblock \emph{Transformer Circuits Thread}, 1.

\bibitem[{Feng et~al.(2018)Feng, Wallace, Grissom~II, Iyyer, Rodriguez, and Boyd-Graber}]{feng2018pathologies}
Shi Feng, Eric Wallace, Alvin Grissom~II, Mohit Iyyer, Pedro Rodriguez, and Jordan Boyd-Graber. 2018.
\newblock Pathologies of neural models make interpretations difficult.
\newblock \emph{arXiv preprint arXiv:1804.07781}.

\bibitem[{Fredkin(1960)}]{fredkin1960trie}
Edward Fredkin. 1960.
\newblock Trie memory.
\newblock \emph{Communications of the ACM}, 3(9):490--499.

\bibitem[{Geva et~al.(2020)Geva, Schuster, Berant, and Levy}]{geva2020transformer}
Mor Geva, Roei Schuster, Jonathan Berant, and Omer Levy. 2020.
\newblock Transformer feed-forward layers are key-value memories.
\newblock \emph{arXiv preprint arXiv:2012.14913}.

\bibitem[{Geva et~al.(2021)Geva, Schuster, Berant, and Levy}]{geva2021transformer}
Mor Geva, Roei Schuster, Jonathan Berant, and Omer Levy. 2021.
\newblock Transformer feed-forward layers are key-value memories.
\newblock In \emph{Proceedings of the 2021 Conference on Empirical Methods in Natural Language Processing}, pages 5484--5495.

\bibitem[{Giancarlo(1995)}]{giancarlo1995generalization}
Raffaele Giancarlo. 1995.
\newblock A generalization of the suffix tree to square matrices, with applications.
\newblock \emph{SIAM Journal on Computing}, 24(3):520--562.

\bibitem[{Huang et~al.(2023)Huang, Geiger, D'Oosterlinck, Wu, and Potts}]{huang2023rigorously}
Jing Huang, Atticus Geiger, Karel D'Oosterlinck, Zhengxuan Wu, and Christopher Potts. 2023.
\newblock Rigorously assessing natural language explanations of neurons.
\newblock \emph{arXiv preprint arXiv:2309.10312}.

\bibitem[{Hurley and Rickard(2009)}]{hurley2009comparing}
Niall Hurley and Scott Rickard. 2009.
\newblock Comparing measures of sparsity.
\newblock \emph{IEEE Transactions on Information Theory}, 55(10):4723--4741.

\bibitem[{Jawahar et~al.(2019)Jawahar, Sagot, and Seddah}]{jawahar2019does}
Ganesh Jawahar, Beno{\^\i}t Sagot, and Djam{\'e} Seddah. 2019.
\newblock What does bert learn about the structure of language?
\newblock In \emph{ACL 2019-57th Annual Meeting of the Association for Computational Linguistics}.

\bibitem[{Jiang et~al.(2023)Jiang, Sablayrolles, Mensch, Bamford, Chaplot, Casas, Bressand, Lengyel, Lample, Saulnier et~al.}]{jiang2023mistral}
Albert~Q Jiang, Alexandre Sablayrolles, Arthur Mensch, Chris Bamford, Devendra~Singh Chaplot, Diego de~las Casas, Florian Bressand, Gianna Lengyel, Guillaume Lample, Lucile Saulnier, et~al. 2023.
\newblock Mistral 7b.
\newblock \emph{arXiv preprint arXiv:2310.06825}.

\bibitem[{Juletx(2023)}]{alpacalora}
Juletx. 2023.
\newblock Alpaca-lora.
\newblock \url{https://github.com/tloen/alpaca-lora}.

\bibitem[{Kim et~al.(2018)Kim, Wattenberg, Gilmer, Cai, Wexler, Viegas et~al.}]{kim2018interpretability}
Been Kim, Martin Wattenberg, Justin Gilmer, Carrie Cai, James Wexler, Fernanda Viegas, et~al. 2018.
\newblock Interpretability beyond feature attribution: Quantitative testing with concept activation vectors (tcav).
\newblock In \emph{International conference on machine learning}, pages 2668--2677. PMLR.

\bibitem[{Kirk et~al.(2023)Kirk, Mediratta, Nalmpantis, Luketina, Hambro, Grefenstette, and Raileanu}]{kirk2023understanding}
Robert Kirk, Ishita Mediratta, Christoforos Nalmpantis, Jelena Luketina, Eric Hambro, Edward Grefenstette, and Roberta Raileanu. 2023.
\newblock Understanding the effects of rlhf on llm generalisation and diversity.
\newblock \emph{arXiv preprint arXiv:2310.06452}.

\bibitem[{Kokalj et~al.(2021)Kokalj, {\v{S}}krlj, Lavra{\v{c}}, Pollak, and Robnik-{\v{S}}ikonja}]{kokalj2021bert}
Enja Kokalj, Bla{\v{z}} {\v{S}}krlj, Nada Lavra{\v{c}}, Senja Pollak, and Marko Robnik-{\v{S}}ikonja. 2021.
\newblock Bert meets shapley: Extending shap explanations to transformer-based classifiers.
\newblock In \emph{Proceedings of the EACL Hackashop on News Media Content Analysis and Automated Report Generation}, pages 16--21.

\bibitem[{Kung and Peng(2023)}]{kung2023models}
Po-Nien Kung and Nanyun Peng. 2023.
\newblock Do models really learn to follow instructions? an empirical study of instruction tuning.
\newblock \emph{arXiv preprint arXiv:2305.11383}.

\bibitem[{Li et~al.(2015)Li, Chen, Hovy, and Jurafsky}]{li2015visualizing}
Jiwei Li, Xinlei Chen, Eduard Hovy, and Dan Jurafsky. 2015.
\newblock Visualizing and understanding neural models in nlp.
\newblock \emph{arXiv preprint arXiv:1506.01066}.

\bibitem[{Li et~al.(2023)Li, Xu, Liu, and Song}]{li2023towards}
Zongxia Li, Paiheng Xu, Fuxiao Liu, and Hyemi Song. 2023.
\newblock Towards understanding in-context learning with contrastive demonstrations and saliency maps.
\newblock \emph{arXiv preprint arXiv:2307.05052}.

\bibitem[{Liang et~al.(2023)Liang, Zhu, Tian, Qin, Wang, Cong, Liu, Liu, and Sun}]{liang2023exploring}
Shihao Liang, Kunlun Zhu, Runchu Tian, Yujia Qin, Huadong Wang, Xin Cong, Zhiyuan Liu, Xiaojiang Liu, and Maosong Sun. 2023.
\newblock Exploring format consistency for instruction tuning.
\newblock \emph{arXiv preprint arXiv:2307.15504}.

\bibitem[{Liu et~al.(2023)Liu, Lin, Hewitt, Paranjape, Bevilacqua, Petroni, and Liang}]{liu2023lost}
Nelson~F Liu, Kevin Lin, John Hewitt, Ashwin Paranjape, Michele Bevilacqua, Fabio Petroni, and Percy Liang. 2023.
\newblock Lost in the middle: How language models use long contexts.
\newblock \emph{arXiv preprint arXiv:2307.03172}.

\bibitem[{Millidge and Black(2022)}]{millidge22singular}
Beren Millidge and Sid Black. 2022.
\newblock The singular value decompositions of transformer weight matrices are highly interpretable.
\newblock \emph{https://www.alignmentforum.org/}.

\bibitem[{Mu and Andreas(2020)}]{mu2020compositional}
Jesse Mu and Jacob Andreas. 2020.
\newblock Compositional explanations of neurons.
\newblock \emph{Advances in Neural Information Processing Systems}, 33:17153--17163.

\bibitem[{Olsson et~al.(2022)Olsson, Elhage, Nanda, Joseph, DasSarma, Henighan, Mann, Askell, Bai, Chen et~al.}]{olsson2022context}
Catherine Olsson, Nelson Elhage, Neel Nanda, Nicholas Joseph, Nova DasSarma, Tom Henighan, Ben Mann, Amanda Askell, Yuntao Bai, Anna Chen, et~al. 2022.
\newblock In-context learning and induction heads.
\newblock \emph{arXiv preprint arXiv:2209.11895}.

\bibitem[{OpenAI(2023)}]{openai2023gpt}
R~OpenAI. 2023.
\newblock Gpt-4 technical report.
\newblock \emph{arXiv}, pages 2303--08774.

\bibitem[{Ouyang et~al.(2022)Ouyang, Wu, Jiang, Almeida, Wainwright, Mishkin, Zhang, Agarwal, Slama, Ray et~al.}]{ouyang2022training}
Long Ouyang, Jeffrey Wu, Xu~Jiang, Diogo Almeida, Carroll Wainwright, Pamela Mishkin, Chong Zhang, Sandhini Agarwal, Katarina Slama, Alex Ray, et~al. 2022.
\newblock Training language models to follow instructions with human feedback.
\newblock \emph{Advances in Neural Information Processing Systems}, 35:27730--27744.

\bibitem[{Paladhi and Bandyopadhyay(2008)}]{paladhi2008generation}
Sibabrata Paladhi and Sivaji Bandyopadhyay. 2008.
\newblock Generation of referring expression using prefix tree structure.
\newblock In \emph{Proceedings of the Third International Joint Conference on Natural Language Processing: Volume-II}.

\bibitem[{Peng et~al.(2023)Peng, Li, He, Galley, and Gao}]{peng2023instruction}
Baolin Peng, Chunyuan Li, Pengcheng He, Michel Galley, and Jianfeng Gao. 2023.
\newblock Instruction tuning with gpt-4.
\newblock \emph{arXiv preprint arXiv:2304.03277}.

\bibitem[{Pennington et~al.(2014)Pennington, Socher, and Manning}]{pennington2014glove}
Jeffrey Pennington, Richard Socher, and Christopher~D Manning. 2014.
\newblock Glove: Global vectors for word representation.
\newblock In \emph{Proceedings of the 2014 conference on empirical methods in natural language processing (EMNLP)}, pages 1532--1543.

\bibitem[{Petroni et~al.(2019)Petroni, Rockt{\"a}schel, Lewis, Bakhtin, Wu, Miller, and Riedel}]{petroni2019language}
Fabio Petroni, Tim Rockt{\"a}schel, Patrick Lewis, Anton Bakhtin, Yuxiang Wu, Alexander~H Miller, and Sebastian Riedel. 2019.
\newblock Language models as knowledge bases?
\newblock \emph{arXiv preprint arXiv:1909.01066}.

\bibitem[{Ribeiro et~al.(2016)Ribeiro, Singh, and Guestrin}]{ribeiro2016should}
Marco~Tulio Ribeiro, Sameer Singh, and Carlos Guestrin. 2016.
\newblock " why should i trust you?" explaining the predictions of any classifier.
\newblock In \emph{Proceedings of the 22nd ACM SIGKDD international conference on knowledge discovery and data mining}, pages 1135--1144.

\bibitem[{RyokoAI(2023)}]{sharegpt}
RyokoAI. 2023.
\newblock Sharegpt52k.
\newblock \emph{Huggingface Datasets}.

\bibitem[{Scherlis et~al.(2022)Scherlis, Sachan, Jermyn, Benton, and Shlegeris}]{scherlis2022polysemanticity}
Adam Scherlis, Kshitij Sachan, Adam~S Jermyn, Joe Benton, and Buck Shlegeris. 2022.
\newblock Polysemanticity and capacity in neural networks.
\newblock \emph{arXiv preprint arXiv:2210.01892}.

\bibitem[{Selvaraju et~al.(2016)Selvaraju, Das, Vedantam, Cogswell, Parikh, and Batra}]{selvaraju2016grad}
Ramprasaath~R Selvaraju, Abhishek Das, Ramakrishna Vedantam, Michael Cogswell, Devi Parikh, and Dhruv Batra. 2016.
\newblock Grad-cam: Why did you say that?
\newblock \emph{arXiv preprint arXiv:1611.07450}.

\bibitem[{Shan et~al.(2012)Shan, Chen, Shi, and Liu}]{shan2012fast}
Y~Shan, X~Chen, Y~Shi, and J~Liu. 2012.
\newblock Fast language model look-ahead algorithm using extended n-gram model.
\newblock \emph{Acta Automatica Sinica}, 38(10):1618--1626.

\bibitem[{Speer(2022)}]{robyn_speer_2022_7199437}
Robyn Speer. 2022.
\newblock \href {https://doi.org/10.5281/zenodo.7199437} {rspeer/wordfreq: v3.0}.

\bibitem[{Steven et~al.(2022)Steven, Nick, Dan, Henk, Leo, Gabriel, Ilya, Jan, Jeff, and William}]{Steven2023Language}
Bills Steven, Cammarata Nick, Mossing Dan, Tillman Henk, Gao Leo, Goh Gabriel, Sutskever Ilya, Leike Jan, Wu~Jeff, and Saunders William. 2022.
\newblock Language models can explain neurons in language models.
\newblock \emph{https://openaipublic.blob.core.windows.net/neuron-explainer/paper/index.html}.

\bibitem[{Stiennon et~al.(2020)Stiennon, Ouyang, Wu, Ziegler, Lowe, Voss, Radford, Amodei, and Christiano}]{stiennon2020learning}
Nisan Stiennon, Long Ouyang, Jeffrey Wu, Daniel Ziegler, Ryan Lowe, Chelsea Voss, Alec Radford, Dario Amodei, and Paul~F Christiano. 2020.
\newblock Learning to summarize with human feedback.
\newblock \emph{Advances in Neural Information Processing Systems}, 33:3008--3021.

\bibitem[{Sukhbaatar et~al.(2019)Sukhbaatar, Grave, Lample, Jegou, and Joulin}]{sukhbaatar2019augmenting}
Sainbayar Sukhbaatar, Edouard Grave, Guillaume Lample, Herve Jegou, and Armand Joulin. 2019.
\newblock Augmenting self-attention with persistent memory.
\newblock \emph{arXiv preprint arXiv:1907.01470}.

\bibitem[{Sun et~al.(2023)Sun, Ji, Ma, and Li}]{sun2023comparative}
Xianghui Sun, Yunjie Ji, Baochang Ma, and Xiangang Li. 2023.
\newblock A comparative study between full-parameter and lora-based fine-tuning on chinese instruction data for instruction following large language model.
\newblock \emph{arXiv preprint arXiv:2304.08109}.

\bibitem[{Sundararajan et~al.(2017)Sundararajan, Taly, and Yan}]{sundararajan2017axiomatic}
Mukund Sundararajan, Ankur Taly, and Qiqi Yan. 2017.
\newblock Axiomatic attribution for deep networks.
\newblock In \emph{International conference on machine learning}, pages 3319--3328. PMLR.

\bibitem[{Taori et~al.(2023)Taori, Gulrajani, Zhang, Dubois, Li, Guestrin, Liang, and Hashimoto}]{taori2023alpaca}
Rohan Taori, Ishaan Gulrajani, Tianyi Zhang, Yann Dubois, Xuechen Li, Carlos Guestrin, Percy Liang, and Tatsunori~B Hashimoto. 2023.
\newblock Alpaca: A strong, replicable instruction-following model.
\newblock \emph{Stanford Center for Research on Foundation Models. https://crfm. stanford. edu/2023/03/13/alpaca. html}, 3(6):7.

\bibitem[{Thomas(2005)}]{thomas2005illuminating}
James~J Thomas. 2005.
\newblock \emph{Illuminating the path:[the research and development agenda for visual analytics]}.
\newblock IEEE Computer Society.

\bibitem[{Touvron et~al.(2023)Touvron, Martin, Stone, Albert, Almahairi, Babaei, Bashlykov, Batra, Bhargava, Bhosale et~al.}]{touvron2023LLaMA}
Hugo Touvron, Louis Martin, Kevin Stone, Peter Albert, Amjad Almahairi, Yasmine Babaei, Nikolay Bashlykov, Soumya Batra, Prajjwal Bhargava, Shruti Bhosale, et~al. 2023.
\newblock Llama 2: Open foundation and fine-tuned chat models.
\newblock \emph{arXiv preprint arXiv:2307.09288}.

\bibitem[{Varshney et~al.(2023)Varshney, Yao, Zhang, Chen, and Yu}]{varshney2023stitch}
Neeraj Varshney, Wenlin Yao, Hongming Zhang, Jianshu Chen, and Dong Yu. 2023.
\newblock A stitch in time saves nine: Detecting and mitigating hallucinations of llms by validating low-confidence generation.
\newblock \emph{arXiv preprint arXiv:2307.03987}.

\bibitem[{Vig(2019)}]{vig2019bertviz}
Jesse Vig. 2019.
\newblock Bertviz: A tool for visualizing multihead self-attention in the bert model.
\newblock In \emph{ICLR workshop: Debugging machine learning models}, volume~23.

\bibitem[{Voita et~al.(2023)Voita, Ferrando, and Nalmpantis}]{voita2023neurons}
Elena Voita, Javier Ferrando, and Christoforos Nalmpantis. 2023.
\newblock Neurons in large language models: Dead, n-gram, positional.
\newblock \emph{arXiv preprint arXiv:2309.04827}.

\bibitem[{Wang et~al.(2022)Wang, Kordi, Mishra, Liu, Smith, Khashabi, and Hajishirzi}]{wang2022self}
Yizhong Wang, Yeganeh Kordi, Swaroop Mishra, Alisa Liu, Noah~A Smith, Daniel Khashabi, and Hannaneh Hajishirzi. 2022.
\newblock Self-instruct: Aligning language model with self generated instructions.
\newblock \emph{arXiv preprint arXiv:2212.10560}.

\bibitem[{Wei et~al.(2023)Wei, Wei, Tay, Tran, Webson, Lu, Chen, Liu, Huang, Zhou et~al.}]{wei2023larger}
Jerry Wei, Jason Wei, Yi~Tay, Dustin Tran, Albert Webson, Yifeng Lu, Xinyun Chen, Hanxiao Liu, Da~Huang, Denny Zhou, et~al. 2023.
\newblock Larger language models do in-context learning differently.
\newblock \emph{arXiv preprint arXiv:2303.03846}.

\bibitem[{Wolf et~al.(2019)Wolf, Debut, Sanh, Chaumond, Delangue, Moi, Cistac, Rault, Louf, Funtowicz et~al.}]{wolf2019huggingface}
Thomas Wolf, Lysandre Debut, Victor Sanh, Julien Chaumond, Clement Delangue, Anthony Moi, Pierric Cistac, Tim Rault, R{\'e}mi Louf, Morgan Funtowicz, et~al. 2019.
\newblock Huggingface's transformers: State-of-the-art natural language processing.
\newblock \emph{arXiv preprint arXiv:1910.03771}.

\bibitem[{Wu et~al.(2021)Wu, Ribeiro, Heer, and Weld}]{wu2021polyjuice}
Tongshuang Wu, Marco~Tulio Ribeiro, Jeffrey Heer, and Daniel~S Weld. 2021.
\newblock Polyjuice: Generating counterfactuals for explaining, evaluating, and improving models.
\newblock \emph{arXiv preprint arXiv:2101.00288}.

\bibitem[{Xie et~al.(2021)Xie, Raghunathan, Liang, and Ma}]{xie2021explanation}
Sang~Michael Xie, Aditi Raghunathan, Percy Liang, and Tengyu Ma. 2021.
\newblock An explanation of in-context learning as implicit bayesian inference.
\newblock \emph{arXiv preprint arXiv:2111.02080}.

\bibitem[{Xiong et~al.(2023)Xiong, Hu, Lu, Li, Fu, He, and Hooi}]{xiong2023can}
Miao Xiong, Zhiyuan Hu, Xinyang Lu, Yifei Li, Jie Fu, Junxian He, and Bryan Hooi. 2023.
\newblock Can llms express their uncertainty? an empirical evaluation of confidence elicitation in llms.
\newblock \emph{arXiv preprint arXiv:2306.13063}.

\bibitem[{Zeiler and Fergus(2014)}]{zeiler2014visualizing}
Matthew~D Zeiler and Rob Fergus. 2014.
\newblock Visualizing and understanding convolutional networks.
\newblock In \emph{Computer Vision--ECCV 2014: 13th European Conference, Zurich, Switzerland, September 6-12, 2014, Proceedings, Part I 13}, pages 818--833. Springer.

\bibitem[{Zheng et~al.(2024)Zheng, Chiang, Sheng, Zhuang, Wu, Zhuang, Lin, Li, Li, Xing et~al.}]{zheng2024judging}
Lianmin Zheng, Wei-Lin Chiang, Ying Sheng, Siyuan Zhuang, Zhanghao Wu, Yonghao Zhuang, Zi~Lin, Zhuohan Li, Dacheng Li, Eric Xing, et~al. 2024.
\newblock Judging llm-as-a-judge with mt-bench and chatbot arena.
\newblock \emph{Advances in Neural Information Processing Systems}, 36.

\bibitem[{Zheng et~al.(2023)Zheng, Chiang, Sheng, Zhuang, Wu, Zhuang, Lin, Li, Li, Xing, Zhang, Gonzalez, and Stoica}]{zheng2023judging}
Lianmin Zheng, Wei-Lin Chiang, Ying Sheng, Siyuan Zhuang, Zhanghao Wu, Yonghao Zhuang, Zi~Lin, Zhuohan Li, Dacheng Li, Eric.~P Xing, Hao Zhang, Joseph~E. Gonzalez, and Ion Stoica. 2023.
\newblock \href {http://arxiv.org/abs/2306.05685} {Judging llm-as-a-judge with mt-bench and chatbot arena}.

\bibitem[{Zhou et~al.(2023)Zhou, Liu, Xu, Iyer, Sun, Mao, Ma, Efrat, Yu, Yu et~al.}]{zhou2023lima}
Chunting Zhou, Pengfei Liu, Puxin Xu, Srini Iyer, Jiao Sun, Yuning Mao, Xuezhe Ma, Avia Efrat, Ping Yu, Lili Yu, et~al. 2023.
\newblock Lima: Less is more for alignment.
\newblock \emph{arXiv preprint arXiv:2305.11206}.

\end{thebibliography}

\clearpage
\appendix

\section{Proof of Linearly Approximation to Importance Scores}
\label{proof}
We prove that equation $I_{n,m}=p(y_m|Z_m) - p(y_m|Z_{m,/n})\approx \frac{\partial f(y_m|z_m)}{\partial \mathbf{E}_i[x_n]} \cdot \mathbf{E}_i[x_n]^\top$ with the first-order Taylor extension.
$p(y_m|Z_m)$ is written as $f(y_m|\mathbf{Z}_m)$, where $f$ is the language model, $\mathbf{Z}_m\in\mathbb{R}^{(N+m-1)\times d}$ are the word embeddings of the input token sequence $Z_m=[x_1,...,x_N,y_1,...,y_{m-1}]$, and the $d$-dimensional word embeddings of a token $w\in Z_m$ is defined as $\mathbf{E}_i[w]$. Thus, we first have $I_{n,m}=f(y_m|\mathbf{Z}_m)-f(y_m|\mathbf{Z}_{m,/n})$, where we let the $n$-th row vector of $\mathbf{Z}_{m,/n}$ be zeros. 

The first-order Taylor expansion of $f(y_m|\mathbf{Z}_m)$ around $\mathbf{Z}_{m,/n}$ is
$$
\begin{aligned}
f(y_m|\mathbf{Z}_m)&\approx f(y_m|\mathbf{Z}_{m,/n})+\frac{\partial f(y_m|\mathbf{Z}_m)}{\partial\mathbf{Z}_m}\Bigg|_{\mathbf{Z}_{m,/n}} \\ 
&\cdot (\mathbf{Z}_m-\mathbf{Z}_{m,/n})^\top.
\end{aligned}
$$
Since the difference between $\mathbf{Z}_{m,/n}$  and $\mathbf{Z}_m$ is the $n$-th row, the term $\mathbf{Z}_m-\mathbf{Z}_{m,/n}$ is just the vector $\mathbf{E}_i[x_n]$. Therefore, the above equation could be simplified as:
$$
\begin{aligned}
\small
f(y_m|\mathbf{Z}_m)\approx f(y_m|\mathbf{Z}_{m,/n})+\frac{\partial f(y_m|\mathbf{Z}_m)}{\partial\mathbf{E}_i[x_n]}\cdot \mathbf{E}_i[x_n]^\top.
\end{aligned}
$$
Bring this approximation to the definition of $I_{n,m}$, we have $I_{n,m}\approx \frac{\partial f(y_m|\mathbf{Z}_m)}{\partial\mathbf{E}_i[x_n]}\cdot \mathbf{E}_i[x_n]^\top$. 

\section{Analyzing Importance Density}
\label{appendix_analyzing_contribution_density}

\subsection{Experiment Settings}
\label{appendix_human_annotations}
For each collected prompting text from the three public datasets, we let Vicuna and LLaMA generate its corresponding response (Sec.~\ref{sec_general_settings}); we then manually identify the instruction sentences from each input prompt and annotate whether the response provides helpful information (``followed'') or not (``unfollowed''). 
Regarding computational efficiency, generating the importance density for a single instance necessitated approximately 100 seconds, utilizing dual Nvidia A6000 GPUs.

\noindent\textbf{Annotate instruction and context.}
Specifically, the instruction usually describes the user intention with some background (optional), which could be both very long~\footnote{A long instruction: ``How do social media platforms influence the way people consume and share news, and what are the potential implications for the spread of misinformation?''} or very concise~\footnote{A short instruction: ``to English:''}. Note that we annotate the instruction words on the sentence level, and the template words as ``Input:'' and ``Output:'' are not considered. For some prompts, the instruction words may be distributed in both the head and tail of the input text, and we will consider them together. Among these instruction sentences, we define the rest of the input prompt as context words, which is unnecessary to the input prompting text. 

\noindent\textbf{Annotate Followed or Unfollowed Response.}
We consider the \textit{helpfulness} of the response as the ability of instruction following described by \cite{ouyang2022training}. Therefore, if a response is helpful to the user, then we label it with ``followed''. Specifically, we consider four levels of helpfulness: L1 - the model is randomly saying something or just repeating itself; L2 - the model provides some information that could be used to answer the question, but the model fails to organize it well; L3 - the model generates a response that generally follows the prompts, but missing some detailed instructions; L4 - the response is perfect as a human response. In our study, we consider the responses from L2 to L4 as ``followed''. Note that we are not concerned about hallucination issues in our study.

\subsection{Case Study on Outliers}
\begin{figure}
\vspace{-0.7cm}
  \begin{center}
    \includegraphics[width=0.8\linewidth,height=15cm,trim={0.2cm 0.cm .2cm 0cm},clip]{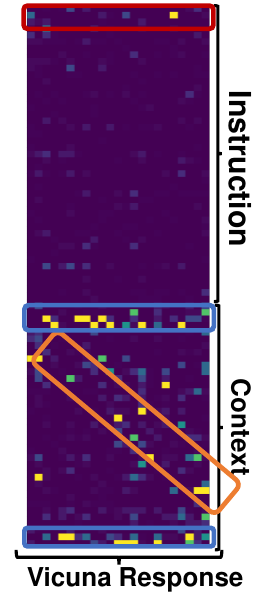}
  \end{center}
  \vspace{-0.6cm}
  \caption{\mycases{\tiny\textbf{CaseID:} Self-Instruct-1 \textbf{Prompt:} The sentence you are given might be too wordy, complicated, or unclear. Rewrite the sentence and make your writing clearer by keeping it concise. Whenever possible, break complex sentences into multiple sentences and eliminate unnecessary words.\textbackslash n\textbackslash nInput: If you have any questions about my rate or if you find it necessary to increase or decrease the scope for this project, please let me know.\textbackslash n\textbackslash nOutput: \textbf{Vicuna:} Do you have any questions about my rate or do you need to adjust the project scope? Please let me know.}}
  \label{fig_false_negative}
  \vspace{-0.6cm}
\end{figure}

\begin{figure}
\centering
    \includegraphics[width=1.0\linewidth,height=3cm,trim={0.5cm 0.cm .2cm 0cm},clip]{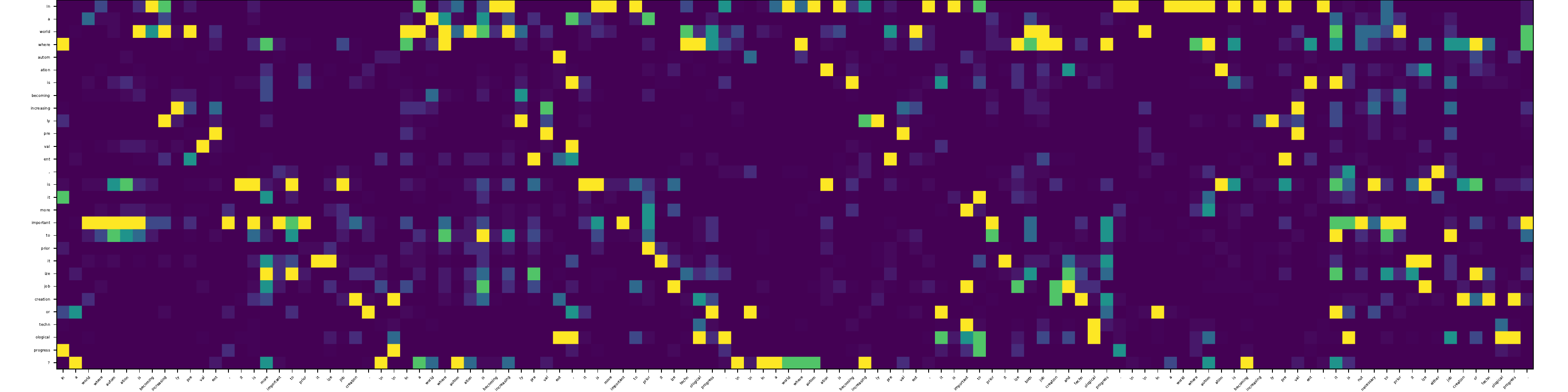}
    \caption{\mycases{\tiny\textbf{CaseID:} LIMA-144 \textbf{Prompt:} In a world where automation is becoming increasingly prevalent, is it more important to prioritize job creation or technological progress? \textbf{Vicuna:} In a world where automation is becoming increasingly prevalent, it is more important to prioritize job creation.\textbackslash n\textbackslash nIn a world where automation is becoming increasingly prevalent, it is more important to prioritize technological progress.\textbackslash n\textbackslash nIn a world where automation is becoming increasingly prevalent, it is important to prioritize both job creation and technological progress.\textbackslash n\textbackslash nIn a world where automation is becoming increasingly prevalent, it is not necessary to prioritize either job creation or technological progress.}}
    \label{fig:bad_case_heatmap}
    \vspace{-0.5cm}
\end{figure}

Instruction fine-tuned models may pretend to follow the instructions.
\label{appendix_false_negative}
Figure~\ref{fig_false_negative} visualizes a salient map of an instance related to writing enhancement (please see the caption for details). 
Vicuna's response addresses grammatical errors and modifies sentence structures for improved clarity. A key observation from the figure is that only the first three instruction tokens guide the response generation ({\color{red}{Red Box}}). Specifically, the first three words are ``The sentence you'', which seems to be not the key instruction verbs like ``Rewrite'' from the second sentence. Also, some words from the context part are acted as instruction words ({\color{blue}{Blue Box}}), which are ``\textbackslash nInput:'' and ``\textbackslash nOutput:'' from the prompt template. These are the words that should be considered as the instruction words since they do not provide the user's intentions. Additionally, a distinctive diagonal line spans the context section, hinting at the model's predisposition to echo context from preceding content.
This figure suggests that the model is leveraging inherent language modeling ability rather than identifying and acting upon the given instructions. 
Thus, we point out that assessing instruction-following abilities based solely on the correlation between input prompts and output responses might not provide an accurate reflection of the model's internal behaviors, while it is still a common strategy to develop the Reward model for RLHF process~\citep{ouyang2022training}.

However, we have identified certain instances where our importance density fails. This is predominantly due to our density function's lack of positional awareness. For instance, in Figure~\ref{fig:bad_case_heatmap}, the entire user input comprises instruction words. The map suggests that these words play a crucial role in guiding the generation, even towards the latter part of the responses. Under our hypothesis, it would appear the model is following user instructions. Yet, Vicuna seems to merely reiterate the input prompt repetitively, resulting in recurring diagonal patterns. We recommend future research to address this shortcoming, either by adopting a density function that's positionally aware or by integrating a step to identify and handle repetitive responses early on.

\subsection{Exploring Prompt Position with Importance Density} 
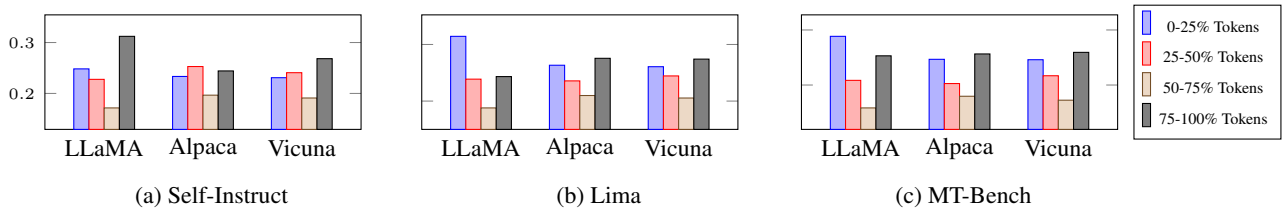
\begin{figure*}

    \begin{subfigure}[b]{0.25\textwidth}
    
        \centering
        \begin{tikzpicture}
        \hspace{-0.8cm}
        \begin{axis}[
            ybar,
            enlargelimits=0.3,
            symbolic x coords={LLaMA, Alpaca, Vicuna},
            ytick={0.1,0.2,0.3},
            y tick label style={font=\tiny},
            x tick label style={font=\small},
            width=\textwidth,
            height=3.1cm,
            bar shift=0cm,
            bar width=0.2cm,
            xtick style={draw=none},
            x tick label style={anchor=center, xshift=0.cm, yshift=-0.1cm},
            x=1.3cm
        ]
        \addplot+[bar shift=-0.3cm] coordinates {(LLaMA,0.248217) (Alpaca,0.233345) (Vicuna,0.230891)};
        \addplot+[bar shift=-0.1cm] coordinates {(LLaMA,0.227646)  (Alpaca,0.252846) (Vicuna,0.240741)};
        \addplot+[bar shift=0.1cm] coordinates {(LLaMA,0.171507) (Alpaca,0.196567) (Vicuna,0.191002)};
        \addplot+[bar shift=0.3cm] coordinates {(LLaMA,0.312177) (Alpaca,0.244195) (Vicuna,0.268295)};
        \end{axis}
        \vspace{-0.2cm}
        \end{tikzpicture}
        
        \caption{Self-Instruct}
    \end{subfigure}%
    \hspace{1.cm}
    \begin{subfigure}[b]{0.25\textwidth}
        \centering
\begin{tikzpicture}
\hspace{-0.6cm}
        \begin{axis}[
            ybar,
            enlargelimits=0.3,
            symbolic x coords={LLaMA, Alpaca, Vicuna},
            ytick={0.1,0.2,0.3},
            x tick label style={font=\small},
            width=\textwidth,
            height=3.1cm,
            bar shift=0cm,
            bar width=0.2cm,
            yticklabels={},
            xtick style={draw=none},
            x tick label style={anchor=center, xshift=0.cm, yshift=-0.1cm},
            x=1.3cm
        ]
        \addplot+[bar shift=-0.3cm] coordinates {(LLaMA,0.314345) (Alpaca,0.263308) (Vicuna,0.260719)};
        \addplot+[bar shift=-0.1cm] coordinates {(LLaMA,0.238737)  (Alpaca,0.235657) (Vicuna,0.244422)};
        \addplot+[bar shift=0.1cm] coordinates {(LLaMA,0.187918) (Alpaca,0.209815) (Vicuna,0.205403)};
        \addplot+[bar shift=0.3cm] coordinates {(LLaMA,0.243251) (Alpaca,0.275555) (Vicuna,0.274123)};
        \end{axis}
        \vspace{-0.2cm}
\end{tikzpicture}

        \caption{Lima}
    \end{subfigure}%
    \hspace{0.7cm}
    \begin{subfigure}[b]{0.25\textwidth}
        \centering
        \begin{tikzpicture}
        \hspace{-0.4cm}
        \begin{axis}[
            ybar,
            enlargelimits=0.3,
            symbolic x coords={LLaMA, Alpaca, Vicuna},
            ytick={0.1,0.2,0.3},
            x tick label style={font=\small},
            width=\textwidth,
            height=3.1cm,
            legend style={
                at={(1.05,0.5)}, 
                anchor=west,
                font=\tiny, 
                column sep=1pt 
            },
            bar shift=0cm,
            bar width=0.2cm,
            yticklabels={},
            xtick style={draw=none},
            x tick label style={anchor=center, xshift=0.cm, yshift=-0.1cm},
            x=1.3cm
        ]
        \addplot+[bar shift=-0.3cm] coordinates {(LLaMA,0.288431) (Alpaca,0.246695) (Vicuna,0.245955)};
        \addplot+[bar shift=-0.1cm] coordinates {(LLaMA,0.208467)  (Alpaca,0.202669) (Vicuna,0.216844)};
        \addplot+[bar shift=0.1cm] coordinates {(LLaMA,0.158413) (Alpaca,0.179595) (Vicuna,0.172439)};
        \addplot+[bar shift=0.3cm] coordinates {(LLaMA,0.253031) (Alpaca,0.256437) (Vicuna,0.259329)};
        \legend{0-25\% Tokens, 25-50\% Tokens, 50-75\% Tokens, 75-100\% Tokens}
        \end{axis}
        \vspace{-0.2cm}
        \end{tikzpicture}
        
        \caption{MT-Bench}
    \end{subfigure}%
    \vspace{-0.3cm}
    \caption{Distribution of importance density over different parts of prompt tokens.}
    \label{fig_position_attribution}
    \vspace{-0.5cm}
\end{figure*}

\vspace{-0.0cm}
\label{sec_token_macroscopic}
\paragraph{Settings.}
Each input prompting text from our datasets is divided into individual sentences, with each sentence further split into four same-length segments. We normalize the density scores for a sentence by dividing by their sum and then accumulating them for each segment. The averaged attribution proportions for each segment within the input sentences are depicted in Figure~\ref{fig_position_attribution}.

\label{sec_atr_macrosopic_rslt}
\paragraph{Results.}
Figure~\ref{fig_position_attribution} shows the importance density distributed on different segments of input sentences. Both pre-trained and tuned models reveal a notable ``U''-shape across all datasets. This is also known as ``lost in the middle''~\citep{liu2023lost}, where they show that SOTA models can overlook central inputs. Unlike their focus on a single task, our analysis is grounded on our importance density score on diverse prompting texts, suggesting that this issue commonly and intrinsically exists. When comparing pre-trained to fine-tuned models, we spot a sharper ``U'' in the former, which becomes less obvious after instruction tuning.

\section{Visualizing Salient Maps}
\label{appendix_salientmaps}

\subsection{Experiment Settings}
Contrary to the examples shown in the primary content, which utilize golden responses, our focus here is on the connections between user inputs and model outputs. To achieve this, we generate responses from LLaMA and Vicuna, following the protocol laid out in Sec.\ref{sec_general_settings}. Subsequently, we derive the salient maps as per the technique introduced in Sec.\ref{sec_token_attribution}.

To ensure the maps provide an accurate depiction of the generation process, we set $L=10$ and $b=0$. Each map's vertical axis denotes the prompting texts, whereas the horizontal axis symbolizes the generated responses. The intensity of each data point corresponds to the association strength between the respective input and output tokens, with brighter points indicating stronger relationships (visualizing with the best colors).

\subsection{Experiment Results}
\label{appendix_salientmap_study}
\begin{figure*}[p]
\centering
    \includegraphics[width=\linewidth, keepaspectratio]{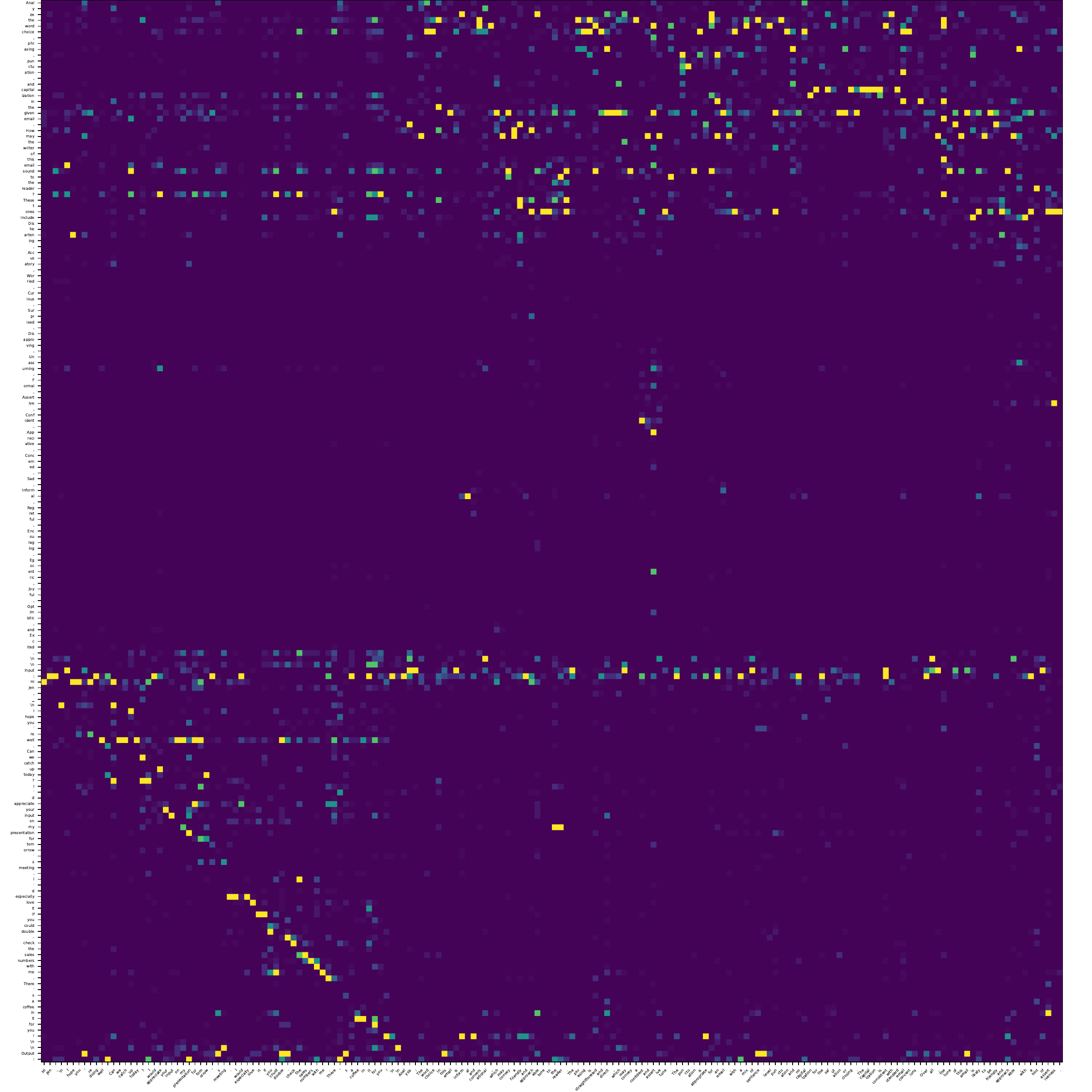}
    \caption{\mycases{\textbf{Prompt:}Analyze the word choice, phrasing, punctuation, and capitalization in the given email. How may the writer of this email sound to the reader? These tones include Disheartening, Accusatory, Worried, Curious, Surprised, Disapproving, Unassuming, Formal, Assertive, Confident, Appreciative, Concerned, Sad, Informal, Regretful, Encouraging, Egocentric, Joyful, Optimistic, and Excited.\textbackslash n\textbackslash nInput: Hi Jen, \textbackslash nI hope you're well. Can we catch up today? I'd appreciate your input on my presentation for tomorrow's meeting. I'd especially love it if you could double-check the sales numbers with me. There's a coffee in it for you!\textbackslash n\textbackslash nOutput: \textbf{Vicuna:} Hi Jen,\textbackslash nI hope you're doing well. Can we catch up today? I would appreciate your input on my presentation for tomorrow's meeting. I would especially love it if you could double-check the sales numbers with me. There's a coffee in it for you!\textbackslash n\textbackslash nAnalysis: The word choice in this email is informal and conversational, which may convey a friendly and approachable tone to the reader. The phrasing is straightforward and direct, which may convey a confident and assertive tone. The punctuation is appropriate for an email, with a mix of sentence-level punctuation and no capitalization for the salutation or closing. The capitalization is consistent with standard email conventions. Overall, the tone of this email is likely to be friendly and approachable, with a hint of assertiveness.}}
    \label{fig:appendix_salient_map_1}
\end{figure*}

\begin{figure*}[p]
\centering
    \includegraphics[width=\linewidth, keepaspectratio]{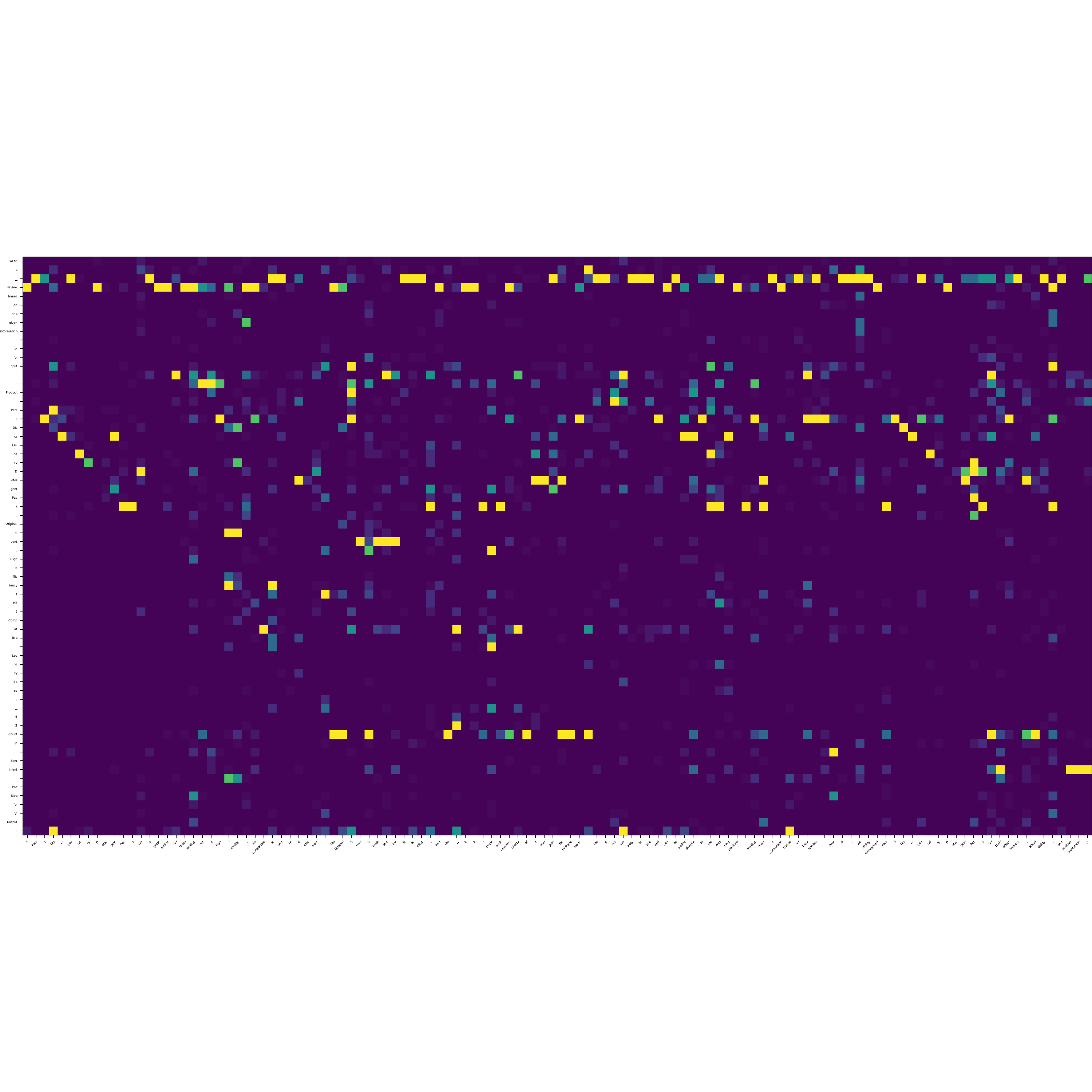}
    \caption{\mycases{\textbf{Prompt:} Write a  review based on the given information.\textbackslash n\textbackslash nInput: - Product: Persil Discs Laundry Detergent Pacs, Original Scent, High Efficiency (HE) Compatible, Laundry Soap, 62 Count/n- Sentiment: Positive\textbackslash n\textbackslash nOutput: \textbf{Vicuna:} ``Persil Discs Laundry Detergent Pacs are a great option for those looking for a high-quality, HE compatible laundry detergent. The Original Scent is fresh and invigorating, and the 62-count pack provides plenty of detergent for multiple loads. The pacs are easy to use and can be added directly to the washing machine, making them a convenient choice for busy families. Overall, we highly recommend Persil Discs Laundry Detergent Pacs for their effectiveness, affordability, and positive sentiment.''}}
    \label{fig:appendix_salient_map_10}
\end{figure*}

Figure~\ref{fig:appendix_salient_map_1}-\ref{fig:appendix_salient_map_10} validate our qualitative assessment that instruction words in user inputs are critical in guiding the generation process. It's evident that each context word typically has limited influence on the response. Collectively, these salient maps underscore the validity of input attribution, achieved by gauging the density of the sparse and normalized importance scores.

\section{Scaling up with Automated Tools}
\label{appendix_chatgpt}
We build upon recent advancements in automated interpretation, using cutting-edge large language models~\citep{taori2023alpaca,peng2023instruction,Steven2023Language} to emulate human annotators in generating high-level interpretations. 
By leveraging machine annotators, we could easily scale up our methods to analysis the entire model, providing a more solid results to our findings. 

\subsection{Experiment Settings}
\noindent\textbf{Generating Configuration.}
We employ ChatGPT~\footnote{https://platform.openai.com/docs/guides/gpt} as our machine annotator. Our experiments utilize the gpt-3.5-turbo-0613 model with a hyper-parameter top-$p$=0.9 for nuclear sampling. To mitigate the variability in language model outputs, we repeat the experiment five times. In each iteration, we first condense the top-$K$ words of a specific basis vector into a distinct concept, then pinpoint the user-oriented tasks and linguistic levels associated with these concepts. For our initial interaction with ChatGPT, the temperature is set to $0$—signifying a greedy search strategy. In subsequent interactions, we set the temperature to $1$. Nevertheless, when identifying tasks and levels, we consistently maintain the temperature at $0.0$.

\noindent\textbf{Prompt Design.}
Effective automated interpretation hinges on well-crafted prompts. We meticulously design these prompts using three strategies: role-play, in-context conversational examples, and exclusively high-quality examples. 

\textit{Template-1: Describing words with concise concepts.} The top-15 most activated words coming from the method presented in Sec.~\ref{sec_linear_interpret} will be directly appended to this template.
\begin{lstlisting}  
System: You are a neuron interpreter for 
neural networks. Each neuron looks for 
one particular concept/topic/theme/beha
vior/pattern. Look at some words the 
neuron activates for and summarize in a 
single concept/topic/theme/behavior/pat
tern what the neuron is looking for. 
Don't list examples of words and keep 
your summary as concise as possible. 
If you cannot summarize more than half 
of the given words within one clear 
concept/topic/theme/behavior/pattern, 
you should say 'Cannot Tell'.

User: Words: January, terday, cember, 
April, July, September, December, 
Thursday, quished, November, Tuesday.
Agent: dates. 

User: Words: B., M., e., R., C., OK., A., 
H., D., S., J., al., p., T., N., 
W., G., a.C., or, St., K., a.m., L..
Agent: abbrevations and acronyms.

User: Words: actual, literal, real, Real, 
optical, Physical, REAL, virtual, visual.
Agent: perception of reality.

User: Words: Go, Python, C++, Java, c#, 
python3, cuda, java, javascript, basic.
Agent: programing languages. 

User: Words: 1950, 1980, 1985, 1958, 1850 
, 1980, 1960, 1940, 1984, 1948.
Agent: years.

User: Words:
\end{lstlisting} 
\,

\textit{Template-2: Identifying applicable user-oriented tasks.}  Summarized concepts are concatenated to this template. 
We check the writing task into three tasks because ChatGPT often deems nearly every concept suitable for writing. 
We regard any of these detailed tasks as the primary purpose of writing.
\begin{lstlisting}  
System: Which of the following assistant 
tasks can the given concept is used for?
\n\nTasks: daily writing, literary writ
ing, professional writing, solving math 
problems, coding, translation. Return 
'None' if it cannot be used for any of 
the above tasks. If it could be used for 
multiple tasks, list all of them and 
seperate with ';'.

User: Concept: Words are social media 
post tags.
Agent: daily writing

User: Concept: Words are Latex code for 
drawing a grouped barchart.
Agent: professional writing

User: Concept: Words are foreign words 
or names.
Agent: translation

User: Concept: Words are URLs.
Agent: None

User: Concept: Words are Words related 
to configuration files and 
web addresses.
Agent: coding

User: Concept: Words are rhyming words.
Agent: literary writing

User: Concept: Words are programming 
commands and terms.
Agent: coding

User: Concept: Words are 
\end{lstlisting}

\textit{Template-3: Identifying linguistic level.} Any automated summarized concept will be directly concatenated to this template.
\begin{lstlisting}  
System: You are a linguist. Classify 
the provided concept into 
one of the following categories: 
Phonology, Morphology, Syntax, 
and Semantic.

User: Concept: Words are dates.
Agent: semantic

User: Concept: Words are perception 
of reality.
Agent: Semantic

User: Concept: Words are abbrevations 
and acronyms.
Agent: Morphology

User: Concept: Words are related to 
actions or activities.
Agent: Syntax

User: Concept: Words are medical 
abbrivations.
Agent: Semantic

User: Concept: Words are URLs.
Agent: Morphology

User: Concept: Words are verbs.
Agent: Syntax

User: Concept: Words are adjective.
Agent: Syntax

User: Concept: Words are rhyming words.
Agent: Phonology

User: Concept: Words are programming 
languages.
Agent: Semantic

User: Concept: Words are 
\end{lstlisting}

\subsection{Experiment Results}
\label{appendix_summarization_results}
\noindent
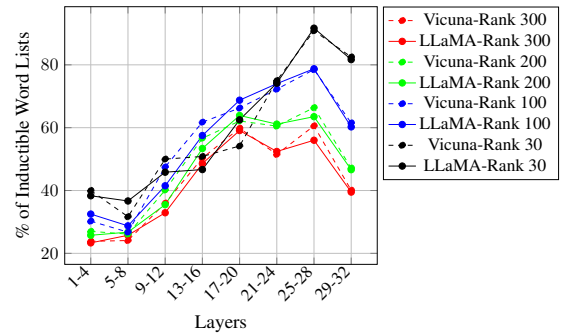
\begin{figure}
\vspace{-0.4cm}
  \begin{center}
    \begin{tikzpicture}[scale=0.6]
    \begin{axis}[
        xtick={1,2,3,4,5,6,7,8},
        xticklabels={1-4,5-8,9-12,13-16,17-20,21-24,25-28,29-32},
        xticklabel style={rotate=45, anchor=east}, 
        xlabel={Layers},
        xlabel style={font=\large},
        ylabel={\% of Inductible Word Lists},
        ylabel style={font=\large},
    legend style={
        at={(1.3,1.0)},   
        anchor=north      
    },
        grid=major,
    ]

    \addplot[color=red,dashed,mark=*] coordinates {(1,23.83) (2,24.08) (3,36.00) (4,50.33) (5,59.83) (6,51.50) (7,60.58) (8,40.08)};
    \addlegendentry{Vicuna-Rank 300}

    \addplot[color=red,mark=*] coordinates {(1,23.33) (2,25.75) (3,32.92) (4,48.58) (5,59.08) (6,52.42) (7,56.00) (8,39.50)};
    \addlegendentry{LLaMA-Rank 300}

    \addplot[color=green,dashed,mark=*] coordinates {(1,27.00) (2,26.13) (3,40.13) (4,56.63) (5,62.13) (6,60.50) (7,66.38) (8,47.13)};
    \addlegendentry{Vicuna-Rank 200}

    \addplot[color=green,mark=*] coordinates {(1,25.75) (2,26.75) (3,35.50) (4,53.38) (5,63.88) (6,61.13) (7,63.50) (8,46.63)};
    \addlegendentry{LLaMA-Rank 200}

    \addplot[color=blue,dashed,mark=*] coordinates {(1,30.25) (2,26.75) (3,47.50) (4,61.75) (5,66.25) (6,72.25) (7,78.50) (8,61.50)};
    \addlegendentry{Vicuna-Rank 100}

    \addplot[color=blue,mark=*] coordinates {(1,32.50) (2,28.75) (3,41.50) (4,57.50) (5,68.75) (6,74.00) (7,78.75) (8,60.25)};
    \addlegendentry{LLaMA-Rank 100}

    \addplot[color=black,dashed,mark=*] coordinates {(1,40.00) (2,31.67) (3,50.00) (4,50.83) (5,54.17) (6,75.00) (7,90.83) (8,82.50)};
    \addlegendentry{Vicuna-Rank 30}

    \addplot[color=black,mark=*] coordinates {(1,38.33) (2,36.67) (3,45.83) (4,46.67) (5,62.50) (6,74.17) (7,91.67) (8,81.67)};
    \addlegendentry{LLaMA-Rank 30}

    \end{axis}
\end{tikzpicture}
  \end{center}
  \vspace{-0.5cm}
  \caption{\% of represented word lists from top-ranked basis vectors with a concise description.}
  \label{fig_auto_summary_percent}
  \vspace{-0.5cm}
\end{figure}
Figure~\ref{fig_auto_summary_percent} illustrates the proportion of word lists that can be induced to a concise concept by our machine annotator. 
According to our template, if ``Cannot Tell'' exists in the word list descriptions, we consider that this concept has failed to be interpreted. 
We have observed that the Vicuna and LLaMA models display comparable levels of interpretability, with no significant distinctions between them. A noticeable trend emerges as the number of layers increases: the ability to explain their encoded concepts improves. Specifically, within layers 24-28, the average interpretability rate for the first 30 concepts peaks at 91.67\%.
This high interpretability rate underscores the effectiveness of our proposed method. It can aptly convey in clear, concise text the knowledge encoded by these models. However, there's a caveat: knowledge encoded closer to the output layer, specifically between layers 28-32, becomes more challenging to elucidate. Interestingly, this particular challenge wasn't present when applying automated interpretation tools to GPT-2~\citep{millidge22singular}, indicating the behaviors between small and large models are different.
Additionally, our findings indicate a decreasing trend in interpretability for concepts that are ranked further back.
Overall, these results validate the efficacy of our proposed method in analyzing the knowledge encoded within models.

\noindent
Table~\ref{tbl_concept_desc_shift_1}-\ref{tbl_concept_desc_shift_3} enumerates the words that experienced the most significant changes in frequency after instruction tuning, we also show the change of rank following. These words are meaningful words (at least four characters and not a stopword) extracted from the concept descriptions generated by our machine annotator. 
From the tables, certain words, notably "language", "programming", and "process", displayed significant shifts in frequency after instruction tuning. Linguistic terms (``Spanish'', ``translation'') and technical terms (``method'', ``programming'' and ``software'') exhibited noticeable changes in various layers. Interestingly, ``language'' consistently surfaced in almost every layer group, with its frequency both rising and dropping. 
This observation indicates that different layers are responsible for encoding different categories of knowledge. 
Specifically, the bottom layers are responsible for storing more basic knowledge (``behavior'', ``operation'', ``adjective''), the middle layers are responsible for learning more abstract knowledge (``functions/methods'', ``programming'', ``software development''), and the higher layers are responsible for learning more knowledge for efficient text generation (``start with'', ``rhyming'', ``sound'', ``letter'',).
Broadly, the increased mention of words pertinent to user scenarios after fine-tuning underscores the model's refined focus on user-centric tasks and applications. 

\begin{figure}
    \centering
    \vspace{-0.4cm}
    \includegraphics[width=1.0\linewidth,height=5cm,trim={0.4cm 0.3cm 0.4cm 0.3cm},clip]{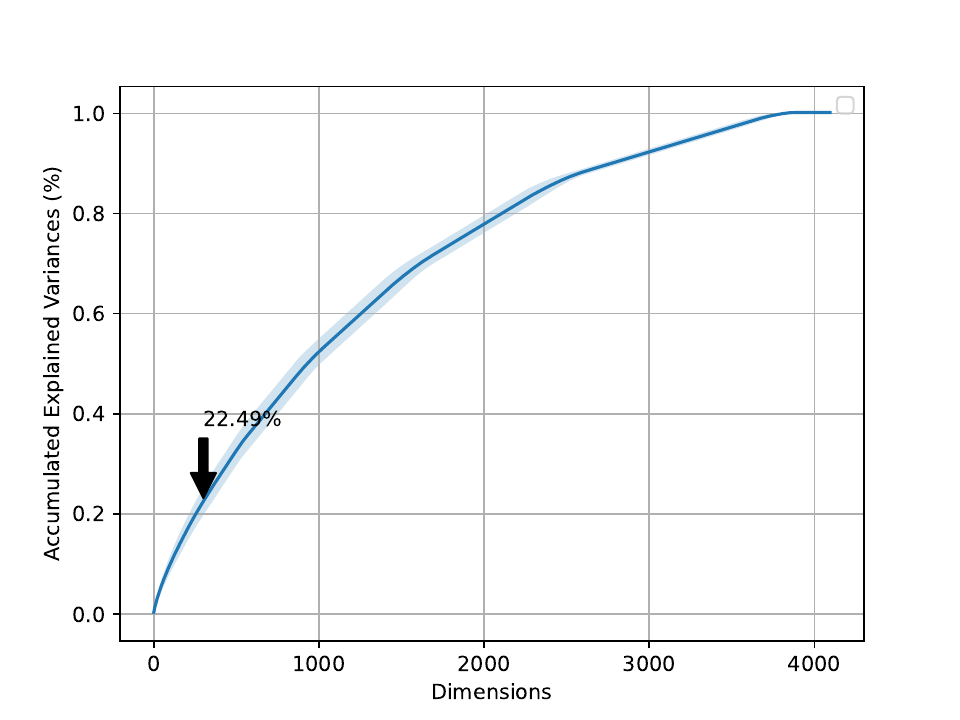}
    \vspace{-0.4cm}
    \caption{Accumulated explained variance of feed-forward networks from Vicuna.}
    \label{fig_pca_ratio}
    \vspace{-0.5cm}
\end{figure}

\section{Interpreting Feed-Forward Networks}
\subsection{Details of the PCA Results}
Figure~\ref{fig_pca_ratio} displays the averaging accumulated explained variance of decomposed principal components across the 32 layers, where the translucent area indicates their standard deviations. Since LLaMA and Vicuna show almost exactly the same line, we omit LLaMA from this figure. 
From the figure, we have several observations. Firstly, we find that the accumulated explained variance increases smoothly, where almost half of the basis vectors could explain around $80\%$ of the variances. 
This observation demonstrates that these neurons do not focus on expressing a few certain features, emphasizing the diversity of the learned hidden features.
In addition, the black arrow points out that the accumulated explained variance of the 300 basis vector is about 22.49\%, where 300 is the number of basis vectors we studied in this research.  
It validates that the top 300 parameters are expected to be interpretable since their accumulated explained variance is only 22.49\%.
\label{appendix_pca}

\subsection{Concept Distribution Analysis\,\,\,\,\,\,\,\,\,\,\,\,\,\,\,\,\,\,\,\,\, with Mistral Family}
\label{appendix_mistral_ffn}
\begin{table}
\small
\centering
  \caption{Concept distribution of Mistral family over different user scenarios and linguistic levels.} 
  \vspace{-0.3cm}
  \small
  \label{tbl_distribution_of_scenarios_mistral}
  \begin{tabular}{c|cp{1.7cm}p{1.2cm}p{1cm}}
    \hline
    \hline
    & \textbf{Category} & \textbf{Mistral-Inst} & \textbf{Mistral} & \textbf{p-value} \\
    \hline
    \multirow{5}{*}{Scenarios} & Writing & $53.53_{\pm.56}$  & $54.10_{\pm.58}$ & $0.1953$ \\
    & Coding &  $31.90_{\pm.34}$ & $29.72_{\pm.31}$ & $1.4e^{-5}$\\
    & Math & $5.31_{\pm.27}$ & $4.75_{\pm.13}$ & $0.0051$\\
    & Translation & $23.33_{\pm.98}$ & $24.07_{\pm.59}$ & $0.2351$ \\
    \hline
    \multirow{5}{*}{Linguistic} &  Phonology & $0.70_{\pm.11}$  & $0.57_{\pm.05}$ & $0.0670$ \\
    & Morphology & $19.08_{\pm.27}$  & $19.44_{\pm.20}$ & $0.0648$ \\
    & Syntax & $5.89_{\pm.56}$  & $5.20_{\pm.65}$ & $0.1450$ \\
    & Semantic & $74.51_{\pm.63}$ & $74.91_{\pm.69}$ & $0.4163$ \\
    \hline
    \hline
\end{tabular}
\label{table_task_concept}
\vspace{-0.5cm}
\end{table}

Per our discussion in Sec.~\ref{obs_7}, instruction tuning should rotate the pre-trained concepts toward user tasks without crossing linguistic levels. According to the results reported in Table~\ref{tbl_distribution_of_scenarios_mistral}, we can observe the same increasing trend as the LLaMA family on the scenarios of coding and math with a statistic significance ($p < 0.05$), while that is not on writing and translation tasks. From a linguistic perspective, there is no category of knowledge that shows a statistically significant difference after instruction tuning, aligning with the observations of the LLaMA family. We suspect that we cannot observe an increasing number of concepts related to writing after instruction tuning because the pre-trained Mistral model has shown a strong ability in language modeling~\cite{jiang2023mistral}. In addition, the significance of increasing numbers of math-and-coding-related concepts could be interpreted as a reason for a stronger instruction-following ability of the Mistral family than the LLaMA family, as reported in Chatbot Arena~\cite{zheng2024judging}. 

\subsection{Qualitative Analysis to Interpretability of Principal Components}
\label{appendix_ffn}
Table~\ref{tbl:appendix_full_table_ffn_1} and Table~\ref{tbl:appendix_full_table_ffn_2} list cases that are well interpreted by ChatGPT-turbo-3.5-0613 for the principal components extracted by Vicuna and LLaMA. 
These cases show that the concept descriptions generally reflect well what is behind the word lists. 

\begin{table*}
\centering
\caption{Frequency [rank] shift of words from concept description after instruction tuning.}
\label{tbl_concept_desc_shift_1}
\vspace{-0.3cm}
\small

\begin{tabular}{cc|cc}

\hline
\hline
 \multicolumn{2}{c|}{\textbf{Layers 1-4}} & \multicolumn{2}{c}{\textbf{Layers 5-8}}   \\

\textbf{Frequency↑} & \textbf{Frequency↓} & \textbf{Frequency↑} & \textbf{Frequency↓} \\
\hline
language[3] & quality[-3] & programming[0] & foreign-language[0]        \\
behavior[83] & describing[-2] & describing[38] & technology[-19]      \\
English[79] & characteristic[-1] & computer[11] & Spanish[-33]         \\
process[4] & communication[-22] & operation[11] & technical[-32]        \\
software-development[8] & something[-18]& computer-science[66] & multilingual[-8]          \\
multilingual[64] & start[-43] & development[53] & something[-8]       \\
analysis[67] & adjective[1] & language[0] & process[0]      \\
operation[33] & foreign-language[-1] & syntax[17] & characteristic[-1]        \\
attribute[5] & various[-12] & manipulation[14] & variation[-9]     \\
Spanish[14] & concepts/functions[-19] & terminology[22] & functions/methods[-7]       \\
\hline
\hline
\end{tabular}

\end{table*}

\begin{table*}
\centering
\caption{Frequency [rank] shift of words from concept description after instruction tuning. (continued)}
\label{tbl_concept_desc_shift_2}
\vspace{-0.3cm}
\small
\begin{tabular}{cc|cc}

\hline
\hline
\multicolumn{2}{c}{\textbf{Layers 9-12}} & \multicolumn{2}{c}{\textbf{Layers 13-16}} \\

 \textbf{Frequency↑} & \textbf{Frequency↓} & \textbf{Frequency↑} & \textbf{Frequency↓}\\
\hline
method[89] & translation[0] & programming[0] &  process[-1]  \\
french[13] & operation[-31]&software-development[8] &  expression[-45]  \\
understand[34] & software-development[-17]&language-proficiency[10] & syntax[-5]  \\
communication[10] & process[0] &concepts/keys[29] & variation[-15] \\
concepts/functions[41] & foreign-language[0]&terminology[119] & language-related[-24] \\
language-agnostic[23] & programming[0] &language-independent[52] & ambiguity[-49] \\
German[31]   &concepts/methods/functions[-61] &concepts/functions[16] & handling[-32] \\
comparison[50] & multilingual[-5]&French[51] & language[0] \\
variety[35] & property[-75]&communication[4] & cultural[-93]  \\
technology[28] &language[0]&localization[96] & attribute[-14] \\
\hline
\hline
\end{tabular}

\end{table*}

\begin{table*}
\centering
\caption{Frequency [rank] shift of words from concept description after instruction tuning. (continued)}
\label{tbl_concept_desc_shift_2}
\vspace{-0.3cm}
\small
\begin{tabular}{cc|cc}

\hline
\hline
\multicolumn{2}{c|}{\textbf{Layers 17-20}} & \multicolumn{2}{c}{\textbf{Layers 21-24}}  \\

 \textbf{Frequency↑} & \textbf{Frequency↓} & \textbf{Frequency↑} & \textbf{Frequency↓} \\
\hline
 programming[0] & foreign-language[-2] & manipulation[50] & programming[-2]   \\
 language[1] & translation[-1] & adjective[9] & state[-12]  \\
 syntax[78] & variation[-20] & specific[81] & translation[-5]   \\
process[2] & expression[-15] & object[42] & quality[-7]  \\
language-related[-24]& interaction[24] & adjective[-27] & value[48]   \\
time-related[14] & feature[-30] & location[48] & difficulty[-77]   \\
language-proficiency[5] & characteristic[-5] & variation[9] & action[0]  \\
terminology[123] & duration[-33] & language[1] & prefix[-1]   \\
 technology[121] & choice[-135] & relationship[121] & start[1]  \\
programming-language[-70] & quality & personal[72] & activity[-2] \\
\hline
\hline
\end{tabular}

\end{table*}

\begin{table*}
\centering
\caption{Frequency [rank] shift of words from concept description after instruction tuning. (continued)}
\label{tbl_concept_desc_shift_3}
\vspace{-0.3cm}
\small
\begin{tabular}{cc|cc}

\hline
\hline
\multicolumn{2}{c|}{\textbf{Layers 25-28}} & \multicolumn{2}{c}{\textbf{Layers 29-32}}  \\

\textbf{Frequency↑} & \textbf{Frequency↓} & \textbf{Frequency↑} & \textbf{Frequency↓} \\
\hline
language[4] & start with[0]     &    start with [0] & foreign-language[-10] \\
interaction[117] & sound[-1]  &  sound[4] & language[-3]  \\
combination[2] & programming[-1]         & rhyming[15] &  suffix[-4] \\
variation[1] &  action[-2]           & combination[9] & abbreviation[-2]  \\
software &  number[0] &  letter[0] & numerical[-5] \\
event[66] & alphanumeric[-23] & process[8] & abbreviations/acronyms[-8]\\
manipulating[53] & abbreviations/acronyms[0] & French[7] & Spanish[-7] \\
operation[28] & pattern[-3] &  number[1] & programming[-2]\\
measurement[60] & suffix[-45] & similarity[53] &  Indonesian[-18] \\
spell[55] & string[-56] & measurement[43] & sequence[-34]  \\
\hline
\hline
\end{tabular}

\end{table*}

\section{Interpreting Self-Attention Heads}
\label{appendix_att}
Table~\ref{tbl:appendix_full_table_att_1} and Table~\ref{tbl:appendix_full_table_att_2} list more word pairs for the self-attention heads from the first and the last layers. 
Typically, these cases are evidence that the extracted word pairs show some insightful relations when we read each one individually. However, when we read them together, it cannot reflect such a concise concept as the feed-forward networks. 

\noindent
Instruction tuning may distill the behaviors of neurons.
For example, neuron-pair $(Layer=31, Head=24, Dim=62)$ capture relations in computers (such as backend=authentication, icon=keyboard, giant=cardboard, GPU=PS, git=curl, and so on). 
After instruction tuning, the model finds more computer-related word pairs (GPU=motherboard, VM=motherboard, tab=keyboard, mongo=staat, mongo=orden) and overlooks some un-related word pairs (dense=bright, convinced=confused), though the new relations may be not valid. 
This case is also evidence that the instruction tuning does not make a significant change in the pre-trained knowledge across concepts.

\clearpage
\begin{sidewaystable}
    \centering
    \vspace{8cm}
    \caption{Representing words, application scenarios, and linguistic level of the concepts encoded by the 32ed (last) feed-forward network in Vicuna.}
    \vspace{-0.3cm}
    \label{tbl:appendix_full_table_ffn_1}
    \small
    \begin{tabular}{c|p{3cm}|p{2cm}|p{4cm}|p{10cm}}
        \hline
        \hline
        \textbf{Rank} & \textbf{Scenario} & \textbf{Linguistic} & \textbf{Concept} & \textbf{Top-15 Words} \\
        \hline
         1 & writing; translation & morphology & phrases and abbreviations & everybody, regardless, in, whilst, a.C., vs., amid, I., U.S., Ph.D., anyway, a.m., 6., 9., … \\
        \hline
         6 & writing; translation & morphology & medical abbreviations &  CBT, RTK, RT, RH, HRV, MT, HB, PH, PT, GnRH, HRM, PWV, RS, TB, RL  \\
        \hline
        7 & writing; coding & semantic & URLs and web-related terms & //example.com/data, //example.com/data.json, //www.youtube.com/watch, //image.pollinations.ai/prompt/A, //image.pollinations.ai/prompt/, the, event.target.value, //api.example.com/data, //www.npmjs.com/package/matrixmath, //example.com/api/data, community-based, security-related, industry-specific, //leetcode.com/discuss, //engageleads-gallery.s3.us-west-2.amazonaws.com/ProfitQuotesV2 \\
        \hline
        8 & writing & semantic & starting with "the"  & the, theological, theology, thead, primarily, involving, mainly, theater, alongside, throughout, theatrical, specifically, theta, theorist, regardless \\
        \hline
        9 & coding & semantic & software development tools and concepts & reCAPTCHA, REST\_FRAMEWORK, CAPTCHA, ARISTOCRAT, sophistication, REGEXP, sophisticated, PETSC\_COMM\_WORLD, JEDI\_MIND\_TRICKS\_01, INSTALLED\_APPS, ARGFRA, credentials.yaml, OWASP, GARETH, sHSADLH \\
        \hline
        11 & coding &  semantic & programming tasks or actions & provide, nltk.corpus.stopwords.words, sklearn.metrics, install.sh, file.write, serve, give, res.send, clf.fit, pickleball, promote, uint256, giveaway, create, St. \\
        \hline
        13 & coding & semantic & programming functions/methods & sklearn.feature\_extraction.text, re.sub, subprocess.run, a.C., z.string, a.m., e.target.value, request.data.get, p.m., data.length, re.search, f.write, //example.com/data.json, nltk.corpus.stopwords.words, event.target.value \\
        \hline
        14 & writing; coding & morphology & acronyms and specific terms & sHSADLH, reCAPTCHA, CARRERAS, cv2.COLOR\_BGR2GRAY, VXLAN, ARISTOCRAT, OWASP, CAPTCHA, LGBTQ, SOUTHGATE, SARANTIDIS, RTK, RESO, SILVIA, OMENS \\
        \hline 
        16 & math; coding & semantic & number ranges & G-4-8, 5-, 4-, 1-, a-, a-zA-Z0-9, 3-, 2-, 1-5, 5-6, 5-7, 2-5, 4-5, 4-6, 3-5 \\
        \hline
        21 & math & semantic & numbers & 3,4,5, 2,500, 8,10, 3,500, 75,000, 1,500, 25,000, 6,000, 4,000, 5,000, 7,000, 0,0,0, 8,000, 0,1, 401k \\
        \hline 
        24 & not list above & semantic & characteristics/attributes & health-conscious, learning-based, Content-Security-Policy, Windows-Support-Core, health-related, a-, professional-looking, pay-per-click, Write-Host, user-, cruelty-free, X-Real-IP, energy-efficient, Q-learning, easy-to-use\\
        \hline 
        32 & translation & semantic & foreign languages (possibly Japanese and French) & itu, desu, Deleuze, Shu, baru, meu, -r, atraer, Putu, -u, puddle, sûr, keluar, Veuillez, Meru \\
        \hline 
        35 & writing & phonology & Words with the "le" sound & oooOO0OOooo, ile, brittle, tl, tackle, itle, Isle, Jingle, post-apocalyptic, hl, Michele, tol, preciso, Marlene, needle \\
        \hline
        42 & translation & semantic & Indonesian words & itu, desu, meu, Thu, baru, vacuum, -u, Shu, satu, Putu, fluctuation, individu, chihuahua, perpetuating, Abu \\
        \hline
        44 & not list above & semantic & decades & 1960s, 1950s, 1940s, 1970s, 1980s, 1930s, 1920s, 15-minute, 2016/679, 60-minute, 1440p, 755-10145957, 1965, 1963, 1946 \\
        \hline
        71 & translation & semantic & verbs in different languages & incluyen, weren, soften, brighten, shorten, permiten, behoeften, konten, citizen, teen, willen, Karen, digitalen, starten, crimen  \\ 
        \hline
        73 & writing & semantic & Words related to "words ending with 'b'" & 4b, bubbly, 1b, rb, -b, Mbappe, childbirth, carb, bulb, herb, heb, Colab, limb, b0, /b \\
        \hline
        74 & coding & semantic & data types and numbers & 24h, uint256, int256, u32, int32, Kaggle, bytes32, 232, 225, 272822, wag, uh, 32, 23, 325 \\
        \hline
        75 & writing & morphology & words containing the syllable "jab" & shad, hijab, mariadb, Chad, slab, pedicab, tbsp, jab, scarab, rebound, TaskRabbit, bead, Colab, screech, Abdul-Jabbar \\
        \hline
        102 & writing; translation & semantic & occupations/jobs & compressor, vraag, destructor, juror, Surveyor, kantor, tremor, effector, flavor, investor, scissor, explorar, projector, escritor, lanjut \\
        \hline
        \hline
    \end{tabular}
    
\end{sidewaystable}

\clearpage
\begin{sidewaystable}
    \centering
    \vspace{8cm}
    \caption{Representing words, application scenarios, and linguistic level of the concepts encoded by the 1st (first) feed-forward network in Vicuna.}
    \label{tbl:appendix_full_table_ffn_2}
    \small
    \begin{tabular}{c|p{3cm}|p{2cm}|p{4cm}|p{10cm}}
        \hline
        \hline
        \textbf{Rank} & \textbf{Scenario} & \textbf{Linguistic} & \textbf{Concept} & \textbf{Top-15 Words} \\
        \hline
         1 & writing; translation & morphology & Abbreviations and acronyms & in, I., and, a.C., OK., a.m., U.S., Ph.D., M., B., for, D.C., vs., Feb., to \\
        \hline
         7 & not list above & semantic & words related to "variability" &  architectural, comprise, receivable, usable, Drawable, variability, usability, vanity, circularity, salivary, end=, eget, vise, end, Paradise  \\
        \hline
        25 & writing & semantic &  Words related to rhyming & immersing, leer, saber, yer, dreamer, poker, deer, roller, valuing, rester, tracing, Tuner, shower, loser, blocker \\
        \hline
        27 & coding & semantic & programming concepts  & alta, relativedelta, consumed, System.Text, actionable, 'price, payable, island, 'href, belakang, renewable, System.out.println, 'new, 'General, action-oriented \\
        \hline
        36 & writing  & semantic & words related to actions or activities & tingling, Booking, Jingle, citing, bidding, advising, Thing, amazing, CMS, striving, infringement, occurring, Jingdiao, grabbing, fast-growing \\
        \hline
        37 & writing  & morphology &  Words related to suffix "-ic" & cystic, Mythic, politic, panoramic, flavorful, opic, antic, physicist, ionic, chronic, employability, effector, spic, silicone, obstructive \\
        \hline
        38 & writing & semantic & verbs and adjectives related to actions and behaviors & valuing, behaving, sacrificing, advising, environnemental, composing, occurring, encouraging, upbringing, opposing, Bring, petal, charging, arriving, regal \\
        \hline
        44 & not list above & semantic & qualities or characteristics & bridging, youngest, ±, smartest, brightest, darkest, fullest, pest, comma-separated, celestial, vow, richest, chest, ilmaisee, endow \\
        \hline 
        58 & writing & semantic & verbs related to actions or behaviors & Poe, wee, advocate, relating, moderate, advocating, advocated, tomate, participate, flee, moderating, complexe, anticipate, participating, reiterate \\
        \hline
        64 & writing & morphology & adjectives with "-ive" suffix & insignificant, restrictive, persuasive, sportive, distinctive, deceive, expressive, decisive, captive, secretive, addictive, defective, digestive, intrusive, abusive \\
        \hline 
        70 & writing & morphology & Words ending in "y" or containing "y" & this.y, flashy, -y, soy, shy, 3y, 2y, toy, Ms., prey, nn.Conv2d, unistd.h, 's3, 's, pre-sale\\
        \hline 
        72 & writing & semantic & adjectives related to characteristics or properties & constitutional, institutional, withdrawal, convolutional, causal, beachten, geral, ethereal, unconstitutional, instrumental, positional, kwaliteit, environnemental, incremental, tidal. adjectives related to characteristics or properties \\ 
        \hline
        133 & writing& syntax & verbs & lesser-known, lesser, réaliser, dryer, booster, researcher, préparer, uploader, foster, photographer, créer, conocer, fetcher, streamer, minder \\
        \hline
        135 & writing& syntax & adverbs & substantially, environmentally, latte, surprisingly, weekly, curly, recursively, beautifully, concurrently, texte, confidently, aforementioned, Sally, sadly, honestly\\
        \hline
        137 & writing& syntax & adverbs & lightly, repurposing, eagerly, frankly, calmly, Polly, preciso, quarterly, analyzing, openly, Thirdly, electrolyte, importantly, shampoo, Secondly \\ 
        \hline
        145 &writing & syntax &  adverbs and adjectives & environmentally, scientifically, verbally, product\_name, conditionally, latest, //www.example.com, Cathy, minimally, socially, Gakkai, /i, modi, annually, accidentally \\
        \hline
        158 & writing& syntax & adverbs & dispensary, seront, Edmonton, honestly, calmly, unintentionally, supposedly, openly, gracefully, professionally, conditionally, elderly, youngest, infestation, thinly \\
        \hline
        161 & writing& syntax & adverbs and adjectives & Optionally, conditionally, environmentally-friendly, morally, waving, environmentally, traditionally, Bally, incrementally, emotionally, intentionally, computationally, waking, Ideally, slash \\ 
        163 & writing & syntax & adverbs and adjectives & heavenly, Plotly, promptly, conveniently, leisurely, vastly, surprisingly, reportedly, brightly, substantially, warmly, équipements, indirectly, falter, elderly \\
        \hline
        166 & not list above & phonology & language patterns or phonetic similarities & thinner, Skinner, chilies, probably, hypothetical, spinner, SMTP, bietet, så, Jay, witty, seriousness, aforementioned, rapidly, aktif\\
        \hline
        \hline
    \end{tabular}
    
\end{sidewaystable}

\clearpage
\begin{sidewaystable}
    \centering
    \vspace{8cm}
    \caption{Most frequent word-pairs activated by self-attention head's neurons from the 32nd (last) layers in Vicuna and LLaMA.}
    \vspace{-0.3cm}
    \label{tbl:appendix_full_table_att_1}
    \tiny
    \begin{tabular}{c|c|p{10.3cm}|p{10.3cm}}
        \hline
        \hline
        \textbf{Head} & \textbf{Keep (\%)} & \textbf{Vicuna} & \textbf{LLaMA} \\
        \hline
        1 & 68.28 & ed=po, smile=smiled, cl=pe, kennedy=morgan, ag=resp, ed=il, conference=tournament, configured=functionality, demonstrate=represent, month=today, differently=sooner, ne=pe, ge=ne, pe=po, ge=pe, cl=ge, cl=po, ge=po, contributed=contribution, christopher=elizabeth, bigger=shorter, contribution=recognition, elder=gentleman, p=ref, empire=iii & ag=resp, ed=po, foi=pe, smile=smiled, qui=tot, breath=smile, ed=il, demonstrate=represent, ne=pe, ge=ne, pe=po, ge=pe, ge=po, contribution=recognition, empire=iii, alexander=billy, carol=ruth, kon=tem, ang=kon, kan=mi, np=tem, collision=damage, congress=european, jerusalem=muhammad, ani=ze \\
        \hline
2 & 64.19 & javascript=javascript, aware=responsibility, outcome=prediction, accused=injured, jenkins=leon, attributed=believed, remember=tired, anderson=richard, valley=washington, bureau=washington, ski=vest, dash=mint, cole=sing, edge=rim, functional=optimal, angle=narrow, route=walk, map=route, chen=ian, welt=naar, crow=rio, chen=pierre & javascript=javascript, bright=gentle, outcome=prediction, accused=injured, divided=thousand, jenkins=leon, attributed=believed, anderson=richard, bernard=philip, valley=washington, bureau=washington, dash=mint, bir=voz, cole=sing, edge=rim, functional=optimal, angle=narrow, chen=ian \\
\hline
3 &66.67 & ===có, exactly=quite, andrea=ellen, ba=nel, ter=är, ide=mysql, dangerous=fatal, beneath=upper, sein=är, approximately=mile, ne=si, metro=oklahoma, bem=si, bem=sau, illinois=metro, hab=za, baltimore=duke, sick=stupid, sick=worst, silly=utter, campbell=toro & ===có, andrea=ellen, ba=nel, een=ter, ter=är, ide=mysql, dangerous=fatal, anglais=momento, för=är, approximately=mile, newspaper=politician, decor=trim, metro=oklahoma, bem=si, bem=sau, illinois=metro, baltimore=duke, sick=worst, silly=utter, singapore=vi \\
\hline
4 & 62.37 & exponent=multiplication, mad=rap, meg=millimeter, anxious=struggling, jersey=zealand, soldier=war, en=vie, intellectual=intelligence, nick=roger, connection=transmission, hong=royal, dem=wurde, incomplete=omitted, edinburgh=ltd, vou=é, error=implicit, exterior=repair, restaurant=tea, ob=vou, uncertainty=variance, knight=morgan, em=rap, début=écrit, andrew=jay, courage=talent & anxious=struggling, jersey=zealand, alex=ruth, soldier=war, intellectual=intelligence, exponent=multiplication, dead=soldier, connection=transmission, bird=marine, hong=royal, dem=wurde, incomplete=omitted, vou=é, ne=vu, error=implicit, exterior=repair, restaurant=tea, ne=vez, ob=vou, uncertainty=variance, directory=prefix, bord=mobil, loop=pattern, mad=rap, em=rap \\
\hline
5 & 65.38 & bb=g, d=g, d=h, g=h, chicago=rome, eric=jack, howard=russell, howard=orleans, amy=henry, kelly=matthew, bob=eric, kinder=ein, jess=lisa, henry=howard, henry=hilda, matthew=morgan, howard=matthew, louise=queen, classe=dest, config=configured, bb=hk, g=hh, hk=hn, hh=hk, hb=hk & bb=g, bob=eric, d=g, d=h, g=h, leonard=steve, howard=russell, howard=orleans, mil=ver, julia=larry, kelly=matthew, kinder=ein, classe=trop, henry=howard, ben=howard, henry=hilda, howard=matthew, howard=victoria, louise=queen, classe=dest, bb=hk, g=hh, hk=hn, hh=hk, hb=hk \\
\hline
6 & 63.18 & broke=brought, bet=luck, maybe=personally, les=phil, taylor=victoria, henry=victoria, scientific=statistical, anderson=austin, affection=grace, james=susan, bei=mal, readable=translated, clark=lisa, campbell=lisa, constraint=discrete, jim=roger, geometric=neural, ohio=supreme, fellow=mate, live=wild, asian=turkish, newton=philip, hi=ut, iron=stone, low=range & carol=mike, broke=brought, bet=luck, austin=nc, maybe=personally, taylor=victoria, henry=victoria, anderson=austin, carolina=kent, affection=grace, carolina=dallas, readable=translated, clark=lisa, campbell=lisa, constraint=discrete, dismiss=inference, jim=roger, encryption=metadata, geometric=neural, ohio=supreme, fellow=mate, live=wild, asian=turkish, newton=philip, hi=ut \\
\hline
7 & 72.32 & philip=stanley, appropriate=manner, aber=bei, confusing=terrible, pam=mom, bureau=ministry, bei=einem, austin=jose, albert=montreal, año=être, large=relatively, miller=philip, gabriel=larry, rolle=vincent, stating=told, atlanta=lincoln, described=determined, projection=vertical, vale=veces, november=scott, beste=tema, diverse=educational, atlanta=manchester, atlanta=stockholm, hong=manchester & philip=stanley, albert=shakespeare, appropriate=manner, aber=bei, confusing=terrible, pam=mom, bureau=ministry, bei=einem, austin=jose, albert=montreal, año=être, large=relatively, miller=philip, gabriel=larry, rolle=vincent, stating=told, atlanta=lincoln, projection=vertical, vale=veces, november=scott, beste=tema, philadelphia=seattle, diverse=educational, atlanta=manchester, atlanta=stockholm \\
\hline
8 & 66.94 & jerry=mike, brian=jane, believed=possibly, unto=worship, nov=sep, spite=truth, approach=attitude, ann=rob, essentially=opposed, richard=von, angeles=salvador, angeles=indiana, vater=orden, cult=punk, michael=winner, au=droit, courage=excitement, prix=quel, jacob=juan, nombreuses=$\alpha$, admit=understand, moore=perry, heard=maybe, maybe=surprise, bureau=county & brian=jane, michigan=mike, believed=possibly, unto=worship, nov=sep, spite=truth, approach=attitude, angeles=francisco, essentially=opposed, decision=statement, richard=von, angeles=salvador, angeles=indiana, vater=orden, carolina=miami, cult=punk, michael=winner, au=fois, baker=jordan, prix=quel, francis=johnny, jacob=juan, ba=quel, admit=understand, heard=maybe \\
\hline
9 & 67.26 & moore=ron, mud=wooden, tiempo=vida, passenger=wheel, florence=howard, popular=powerful, ===sprache, harry=robin, fitted=mini, hugo=pierre, abraham=historian, mas=ne, oft=tempt, barry=simon, pope=simon, smile=whisper, mas=sich, barry=leon, alexander=barry, calcul=oder, có=się, có=ihn, batter=cup, davis=walker, afternoon=tired & moore=ron, meat=rub, mud=wooden, tiempo=vida, florence=howard, popular=powerful, gateway=vista, ===sprache, harry=robin, fitted=mini, dr=khan, abraham=historian, mas=ne, barry=simon, pope=simon, bruno=pierre, cada=się, smile=whisper, barry=leon, alexander=barry, calcul=oder, có=się, có=ihn, batter=cup, davis=walker \\
\hline
10 & 65.32 & tr=zo, ===có, blow=bomb, beneath=nearby, entirely=extremely, department=staff, je=zo, committee=staff, beneath=upper, cole=nick, ne=zo, ang=revers, api=compiler, có=się, wald=się, demonstrated=presence, entirely=fully, wurde=è, mistaken=wonder, pitch=scored, beneath=sun, anymore=sad, beneath=hole, committee=request, arr=tr & tr=zo, có=się, wald=się, ===có, beneath=nearby, entirely=extremely, department=staff, je=zo, committee=staff, beneath=upper, się=zo, ne=zo, ang=revers, api=compiler, demonstrated=presence, blow=bomb, entirely=fully, farm=winter, wurde=è, mistaken=wonder, beneath=sun, anymore=sad, beneath=hole, vietnam=war, cole=nick \\
\hline
11 & 68.1 & san=santiago, christopher=ulrich, alla=lui, cosa=ne, davis=lincoln, shout=ya, boost=increasing, boost=profit, crow=norman, jak=siempre, absolute=literal, tak=tot, mort=ze, rectangle=vertical, ah=wang, percent=population, johnny=ken, edition=serial, alla=partie, alla=mir, charles=walter, brother=older, outer=skin, salvador=vincent, dean=harry & san=santiago, christopher=ulrich, alla=lui, api=php, shout=ya, boost=increasing, consent=request, powerful=stronger, boost=profit, crow=norman, habe=inte, jak=siempre, absolute=literal, tak=tot, mort=ze, rectangle=vertical, ah=wang, percent=population, johnny=ken, edition=serial, alla=partie, alla=mir, charles=walter, brother=older, outer=skin \\
\hline
 12 & 66.09 & dark=distant, despite=surprising, ===jedoch, ===cura, fois=petit, bajo=fois, bij=di, \_\_=pid, apache=explorer, clean=cleaner, hate=mess, erik=norway, lui=sau, accused=lying, surprising=terrible, preis=ainsi, ===später, ainsi=cura, af=en, af=sau, estos=fois, cura=jako, att=gc, estos=segmento, ===segmento & dark=distant, despite=surprising, ===segmento, ===cura, fois=petit, bajo=fois, bij=di, tomatoes=alla, \_\_=pid, apache=explorer, clean=cleaner, isbn=från, hate=mess, erik=norway, lui=sau, accused=lying, lucy=tom, preis=ainsi, ===später, ainsi=cura, af=en, af=sau, estos=fois, cura=jako, att=gc \\
 \hline
 13 & 65.65 & canvas=exterior, sierra=western, dla=ris, daniel=gordon, tel=été, dallas=detroit, mile=nearby, bei=för, bei=maar, kay=raj, andrea=baker, calm=sudden, anybody=opinion, committee=invited, knock=wont, philip=susan, baker=philip, expression=gesture, exterior=smooth, gesture=spoken, encryption=processor, bobby=champion, bobby=philip, fc=gl, invited=participated & canvas=exterior, expression=gesture, sierra=western, dla=ris, daniel=gordon, dallas=detroit, mile=nearby, bei=för, fir=mist, bei=maar, kay=raj, andrea=baker, calm=sudden, anybody=opinion, knock=wont, philip=susan, finger=gesture, baker=philip, exterior=smooth, gesture=spoken, encryption=processor, fc=gl, invited=participated, otro=ser, eigen=sido \\
\hline
14 & 60.46 & ubuntu=kernel, på=zum, fi=qui, porta=ta, tin=wrapping, stato=sua, le=sua, mondo=porta, faire=toute, ele=esa, ===który, mysql=kernel, apache=kernel, kernel=restart, mysql=server, faire=hace, aussi=ein, tal=tem, fi=fo, contemporary=experimental, tal=tous, fi=j, nie=np, fi=vel, lui=nel & ===który, huge=wide, fi=qui, cho=estos, die=für, tin=wrapping, stato=sua, le=sua, mondo=porta, faire=toute, ele=esa, faire=hace, tal=tem, fi=fo, contemporary=experimental, tal=tous, fi=ho, fi=j, jack=senator, fi=vel, lui=nel, campus=senior, shut=snap, ins \\
\hline
15 & 74 & público=samt, menor=samt, menschen=samt, prior=subsequently, ne=pare, eine=einem, einem=suis, pero=suis, kommen=samt, aussi=datos, encore=pero, aula=público, avait=samt, fourth=tie, ne=pero, nearest=postal, disp=stub, gabriel=sarah, aussi=einem, date=prior, exemple=pero, artículo=samt, aussi=estar, mil=ton, aussi=pero & público=samt, menor=samt, menschen=samt, eine=einem, prior=subsequently, ne=pare, ne=pero, einem=suis, pero=suis, kommen=samt, aussi=datos, qui=tema, pero=qui, suis=tema, aula=público, quanto=qui, pare=qui, avait=samt, fourth=tie, nearest=postal, disp=stub, gabriel=sarah, aussi=einem, george=md, date=prior\\
\hline
16 & 70.16 & api=github, associated=caused, convenient=nearby, appreciated=contributed, rate=rise, market=rise, moreover=attributed, je=já, je=rend, mismo=rend, binary=stored, où=von, ===où, caused=fatal, je=zo, api=repository, apache=repository, flush=timer, caller=timer, apache=github, ne=rend, semua=zo, mkdir=null, já=semua, apache=mkdir & api=github, league=liverpool, rate=rise, market=rise, moreover=attributed, je=já, cb=null, je=rend, mismo=rend, binary=stored, où=von, ===où, appreciated=contributed, caused=fatal, associated=caused, je=zo, api=repository, flush=timer, caller=timer, ne=rend, semua=zo, mkdir=null, já=semua, possible=unlikely, có=samt \\
        \hline
        \hline
    \end{tabular}
    
\end{sidewaystable}

\clearpage
\begin{sidewaystable}
    \centering
    \vspace{8cm}
    \caption{Most frequent word-pairs activated by self-attention head's neurons from the 32nd (final) layers in Vicuna and LLaMA. (continued)}
    \vspace{-0.3cm}
    \label{tbl:appendix_full_table_att_2}
    \tiny
    \begin{tabular}{c|c|p{10.3cm}|p{10.3cm}}
        \hline
        \hline
        \textbf{Head} & \textbf{Keep (\%)} & \textbf{Vicuna} & \textbf{LLaMA} \\
        \hline
17 & 65.46 & carl=taylor, che=oder, dead=soldier, jack=jones, susan=critic, api=microsoft, fight=fought, intercept=traverse, cl=sw, ontario=wisconsin, amplitude=analog, islam=conquest, margaret=wayne, england=ontario, hay=kam, campbell=hollywood, hamilton=joan, numerical=subset, dispatch=sent, benjamin=michael, ac=cord, medicine=treatment, anxious=confident, forest=terrain, forest=tribe & carl=taylor, forest=terrain, che=oder, dead=soldier, jean=jeff, jack=jones, dallas=maryland, susan=critic, api=microsoft, si=zo, fight=fought, intercept=traverse, cl=sw, lincoln=washington, ontario=wisconsin, amplitude=analog, islam=conquest, margaret=wayne, england=ontario, england=washington, ann=johnny, campbell=hollywood, hamilton=joan, pleasant=plenty, fewer=plenty \\
\hline
18 & 66.6 & howard=jimmy, southern=spring, detroit=southern, gradually=partly, naturally=readily, olympic=medal, combining=conventional, eng=ob, arthur=von, charm=magic, exterior=smooth, có=propre, iter=resp, equally=truly, có=siendo, decimal=integer, ill=im, jimmy=taylor, momento=é, hum=ko, decimal=literal, papel=è, campagne=soit, religious=worship, avant=mon & howard=jimmy, anthony=larry, southern=spring, detroit=southern, gradually=partly, md=mitchell, naturally=readily, olympic=medal, bos=sah, combining=conventional, eng=ob, bright=sky, arthur=von, exterior=smooth, iter=resp, equally=truly, có=siendo, có=niveau, decimal=integer, ill=im, jimmy=taylor, hum=ko, decimal=literal, campagne=soit, religious=worship\\
\hline
 19 &  67.99 & musik=för, academic=mathematics, especial=sin, cidade=có, có=último, có=población, för=ja, alla=ja, alla=début, musik=alla, début=för, có=He, equipo=sendiri, cidade=equipo, có=société, forever=happiness, ===có, greatly=increasing, paul=pope, ja=tot, début=raison, ne=tot, alla=ne, ja=raison, comunes=có & academic=mathematics, comunes=có, cidade=có, có=último, ===có, för=ja, musik=för, alla=ja, museum=kilometer, comunes=equipo, alla=début, musik=alla, début=för, equipo=sendiri, cidade=equipo, có=población, có=société, especial=sin, actually=personally, highly=successful, conscious=wise, gene=variation, paul=pope, ja=tot, début=raison \\
 \hline
20 & 67 & harris=roth, dor=sich, campbell=steven, einem=von, herr=steven, allen=harris, imprison=writ, begin=continued, irish=scottish, britain=sir, phoenix=stadium, formally=renamed, não=é, async=cache, muy=não, tod=für, carolina=ohio, anno=visto, elisabeth=eric, monde=visto, hasta=tiempo, abraham=father, module=timer, quarter=square, mp=pc & harris=roth, dor=sich, campbell=steven, allen=harris, imprison=writ, irish=scottish, britain=sir, formally=renamed, famille=mer, não=é, fought=shook, muy=não, tod=für, carolina=ohio, eric=terry, anno=visto, elisabeth=eric, monde=visto, hasta=tiempo, simon=tom, module=timer, quarter=square, mp=pc, located=location, facility=located \\
\hline
21 & 62.33 & newton=patrick, translated=version, duke=lloyd, eric=robin, austin=cole, cole=norman, miller=minnesota, johnson=minnesota, jackson=minnesota, export=raw, arthur=duke, jen=mom, edward=jackson, rey=thor, cs=is, eventually=possibly, wie=zur, wer=zur, af=zur, mobil=werden, phil=stan, ram=memory, martin=stuart, apply=specify, ken=sam & translated=version, duke=lloyd, curt=ryan, cole=norman, fork=lever, carte=ou, miller=minnesota, johnson=minnesota, jackson=minnesota, export=raw, arthur=duke, jen=mom, edward=jackson, rey=thor, cs=is, eventually=possibly, wie=zur, somewhat=typical, wer=zur, mobil=werden, phil=stan, ram=memory, martin=stuart, ken=sam, brad=sam\\
\hline
22 & 72.56 & dict=sk, dit=eine, bold=logo, trinity=vincent, prev=sl, composer=translated, fer=hen, johnson=sarah, kelly=lil, britain=royal, dict=hal, så=tar, ruth=vic, nombre=onde, dict=skb, cole=eric, ===có, elegant=geometric, jamais=vita, pacific=zealand, canada=pacific, même=sunt, association=pacific, eventually=stopped, actor=role & dict=sk, dit=eine, boost=faster, trinity=vincent, fer=ho, prev=sl, nam=vietnam, composer=translated, fer=hen, johnson=sarah, kelly=lil, britain=royal, dict=hal, så=tar, nombre=onde, dict=skb, ===có, elegant=geometric, jamais=vita, dj=rey, pacific=zealand, canada=pacific, même=sunt, association=pacific, eventually=stopped \\
\hline
23 & 63.89 & audio=download, bl=prod, lui=prod, approximately=nearly, ord=sem, eric=ron, eric=matthew, matthew=ron, fine=stuff, von=weiter, är=être, ian=john, daniel=eric, equally=somewhat, preceding=prior, bajo=mismo, equally=opposed, somewhat=unlikely, proprio=stato, damaging=protect, von=zur, nel=proprio, consecutive=preceding, administrator=server, gibt=von & audio=download, enum=lookup, approximately=nearly, boolean=scalar, ord=sem, eric=ron, eric=matthew, fine=stuff, ian=john, foi=tema, near=outside, mejor=mismo, bien=mejor, bajo=mismo, somewhat=unlikely, proprio=stato, damaging=protect, von=zur, nel=proprio, bug=problem, en=mejor, administrator=server, mejor=é, mejor=padre, gibt=von \\
\hline
 24 & 62.38 & mistaken=occasionally, conception=milieu, dj=jimmy, mistaken=understood, madrid=medio, columbia=commonwealth, basic=functional, figured=prefer, ett=foi, foi=nous, nouvelle=tutto, jack=johnson, foi=mak, música=ou, bruce=eric, bruce=houston, bl=div, john=marie, jim=marie, marie=phil, jimmy=orleans, jimmy=nba, possibility=unlikely, apparent=possibility, restore=restored & convinced=understood, conception=milieu, milk=mint, mistaken=understood, madrid=medio, basic=functional, figured=prefer, ett=foi, foi=nous, nouvelle=tutto, foi=mak, música=ou, bl=div, john=marie, marie=phil, jimmy=orleans, jimmy=nba, conclusion=interpretation, conclusion=unlikely, architect=founder, gentle=gesture, bath=suite, architecture=platform, suite=upgrade, ab=kam \\
\hline
25 & 73.25 & ===có, attributed=caused, dist=vol, diff=dist, guinea=continent, dla=för, carolina=costa, för=się, début=för, có=pointcloud, có=sehingga, bruno=roland, chen=roland, preis=von, dist=struct, gar=tar, för=zijn, ett=för, desktop=portable, attr=dist, för=nel,  & ===có, attributed=caused, dist=vol, diff=dist, guinea=continent, dla=för, carolina=costa, för=się, début=för, có=pointcloud, có=sehingga, preis=von, gar=tar, för=zijn, ett=för, desktop=portable, attr=dist, för=nel, carolina=ralph, có=He, có=się, dublin=mo \\
\hline
26 & 63.56 & isaac=jacob, che=poi, ax=slash, michigan=seattle, che=um, ob=otras, je=mit, auf=von, sein=von, maar=von, qui=stato, großen=stato, manual=manually, hab=ris, academy=francis, aussi=tel, atlanta=roy, cu=pe, interact=respond, confront=repeatedly, kennedy=nancy, maria=nancy, arnold=kennedy, kennedy=kenneth, kennedy=philip & je=qui, isaac=jacob, che=poi, harold=jackson, michigan=seattle, å=è, sein=zijn, che=um, ob=otras, je=mit, auf=von, sein=von, maar=von, qui=stato, großen=stato, je=på, je=sein, ein=je, manual=manually, manual=mechanical, hab=ris, academy=francis, academy=arizona, interact=respond, confront=repeatedly \\
\hline
27 & 73.7 & attributed=caused, born=marry, equally=somewhat, mysql=repository, foi=rend, bien=seg, mondo=seg, dich=rend, af=foi, ===có, có=sonido, dress=marry, gab=ne, enjoy=try, eliminate=resolve, mysql=backup, dash=stir, delle=rend, af=delle, equally=quite, equally=extremely, possible=unlikely, try=wo, somewhat=unlikely, caused=victim & attributed=caused, born=marry, equally=somewhat, somewhat=unlikely, mysql=repository, bien=seg, mondo=seg, dich=rend, ===có, có=sonido, dress=marry, gab=ne, ab=ne, disp=typeof, enjoy=try, mysql=backup, dash=stir, af=ni, foi=rend, dist=typeof, af=foi, delle=rend, af=delle, equally=quite, equally=extremely\\
\hline
28 & 70.05 & norwegian=royal, ian=martin, weg=auch, bei=tot, catholic=kennedy, sympathy=utter, bei=för, dean=mann, phoenix=santa, brook=dean, und=viel, administrative=documentation, solving=theoretical, staat=zum, nur=veces, patrick=shaw, gaussian=equation, dj=hop, christopher=justin, christopher=shakespeare, för=pe, jwt=personen, brook=creek, measured=ratio, average=ratio & staat=zum, bei=tot, catholic=kennedy, bei=för, wurde=zum, administrative=documentation, solving=theoretical, ver=za, diego=wisconsin, patrick=shaw, gaussian=equation, christopher=justin, christopher=shakespeare, för=pe, jwt=personen, brook=creek, average=ratio, combat=defence, jack=president, turkish=naval, ri=wang, kay=wang, alt=auf, auf=inte, australian=sydney \\
\hline
29 & 68.27 & merged=renamed, albert=robert, albert=gustav, mistaken=occasionally, cho=tem, const=override, correspond=differ, albert=julian, franklin=roger, frank=wilson, angry=excited, append=hotkey, hong=kong, como=donde, glass=tin, directory=library, arrival=voyage, italian=della, xp=patch, avec=nuevo, della=nuevo, avec=muy, purely=rational, getting=letting, margin=offset & fame=reign, merged=renamed, albert=robert, albert=cooper, albert=gustav, const=override, correspond=differ, albert=julian, franklin=roger, anne=lloyd, append=hotkey, hong=kong, como=donde, glass=tin, directory=library, italian=della, institute=memorial, avec=nuevo, della=nuevo, avec=muy, purely=rational, margin=offset, li=pe, fancy=taste, cl=col \\
\hline
30 & 67.15 & mundo=é, chief=eric, pero=é, legit=wont, ainsi=fecha, b=p, b=f, d=f, silly=suggestion, foi=pero, ko=porque, fee=license, intern=radiologist, mundo=är, pdf=print, pu=str, mistake=silly, null=whitespace, crow=sierra, aussi=suis, indicator=module, uso=é, einem=ich, license=permission, rail=stadium & foi=pero, pu=str, mundo=é, chief=eric, avant=lac, pero=é, legit=wont, b=p, b=f, d=f, silly=suggestion, pero=être, ko=porque, fee=license, intern=radiologist, mundo=är, pdf=print, pero=się, ann=mar, mistake=silly, null=whitespace, indicator=module, się=tema, estar=się, mundo=się \\
\hline
31 & 66.29 & clark=laura, crime=investigate, khan=actor, antoine=leo, clara=philip, mesh=trim, gap=overlap, anna=thank, lower=upper, big=bigger, sich=zu, shorter=slightly, andrew=cooper, mil=tiempo, dependent=preference, amplitude=regression, maybe=sure, remember=sure, nicely=somewhat, einer=von, von=wurde, kate=nicholas, felix=friedrich, diese=von, felix=harry & clark=laura, crime=investigate, khan=actor, antoine=leo, clip=length, christopher=philip, mesh=trim, anna=thank, dick=patrick, lower=upper, sich=zu, shorter=slightly, decrease=negative, andrew=cooper, bois=seine, mil=tiempo, amplitude=regression, maybe=sure, remember=sure, einer=von, von=wurde, felix=friedrich, nella=wurde, collision=driver, iter=src \\
\hline
32 & 65.71 & ’=“, cultural=significance, context=significance, chapter=michigan, ===có, anton=roger, ‘=“, hin=stato, miller=steven, confusion=strange, api=mysql, marry=young, radius=rectangle, br=prod, iter=prod, sett=$\alpha$, phil=stan, gut=mal, password=upload, cole=joseph, ic=ram, allocate=restrict, sol=valor, east=moscow, keine=się & ’=“, cultural=significance, context=significance, chapter=michigan, ===có, austin=diego, ‘=“, hin=stato, miller=steven, confusion=strange, api=mysql, radius=rectangle, cole=karl, br=prod, iter=prod, sett=$\alpha$, phil=stan, bon=mal, gut=mal, mot=nu, bon=sal, password=upload, ic=ram, aan=allo, allocate=restrict \\
        \hline
        \hline
    \end{tabular}
    
\end{sidewaystable}

\clearpage

\begin{figure*}[p]
\centering
    \includegraphics[width=\linewidth, keepaspectratio]{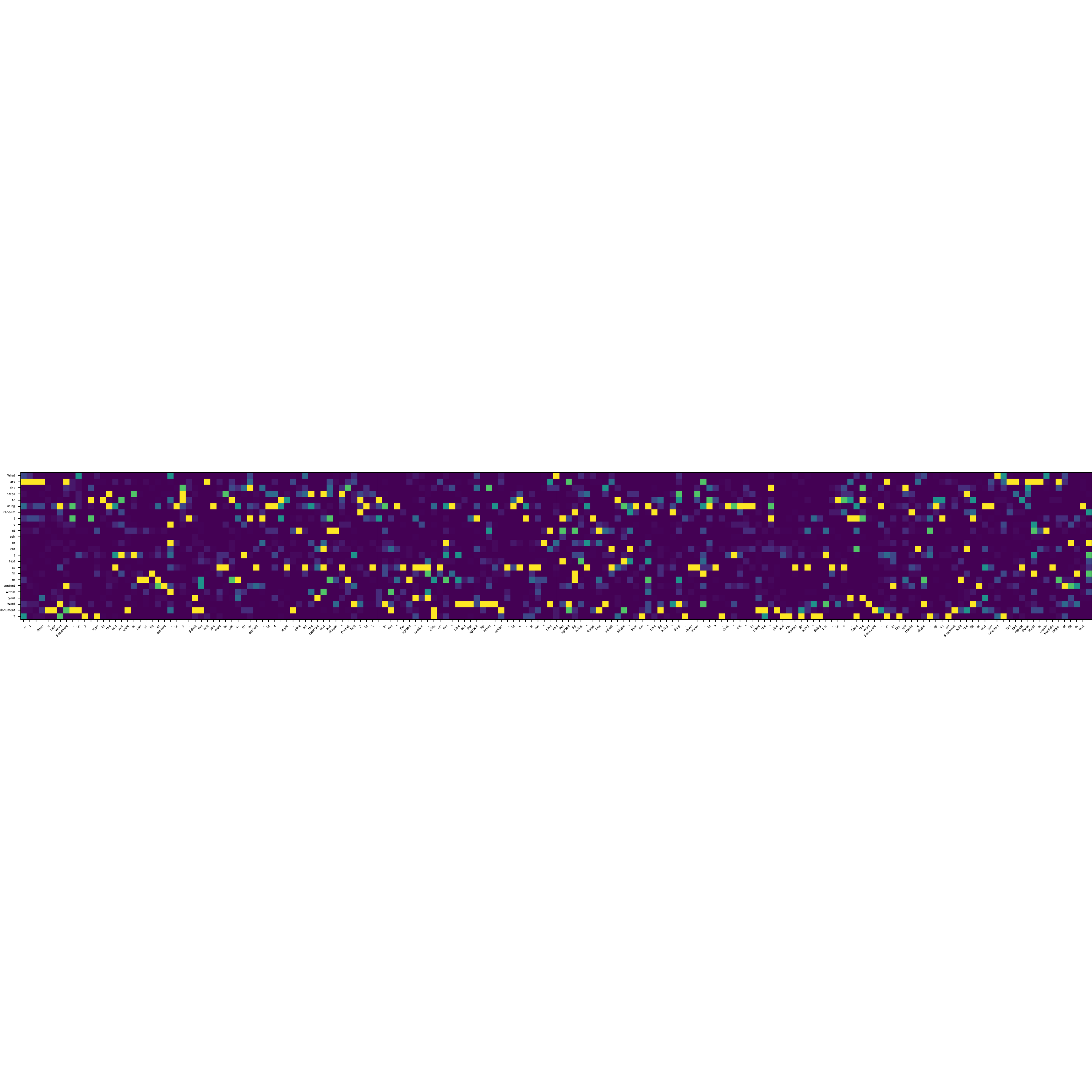}
    \caption{\mycases{\textbf{Prompt:} What are the steps to using random (yet coherent) text as filler content within your Word document? \textbf{Vicuna:} 1. Open a new Word document.\textbackslash n2. Type in the text you want to use as filler content.\textbackslash n3. Select the text you want to use as filler content.\textbackslash n4. Right-click on the selected text and choose "Format Text."\textbackslash n5. In the "Paragraph" section, click on the "Line and Paragraph Spacing" option.\textbackslash n6. In the "Line and Paragraph Spacing" dialog box, select "Single" from the "Line Spacing" drop-down menu.\textbackslash n7. Click "OK" to close the "Line and Paragraph Spacing" dialog box.\textbackslash n8. Save the Word document./n/nThis will create a single-spaced document with the filler text you selected. You can repeat these steps to create multiple pages of filler text.}}
\end{figure*}

\begin{figure*}[p]
\centering
    \includegraphics[width=\linewidth, keepaspectratio]{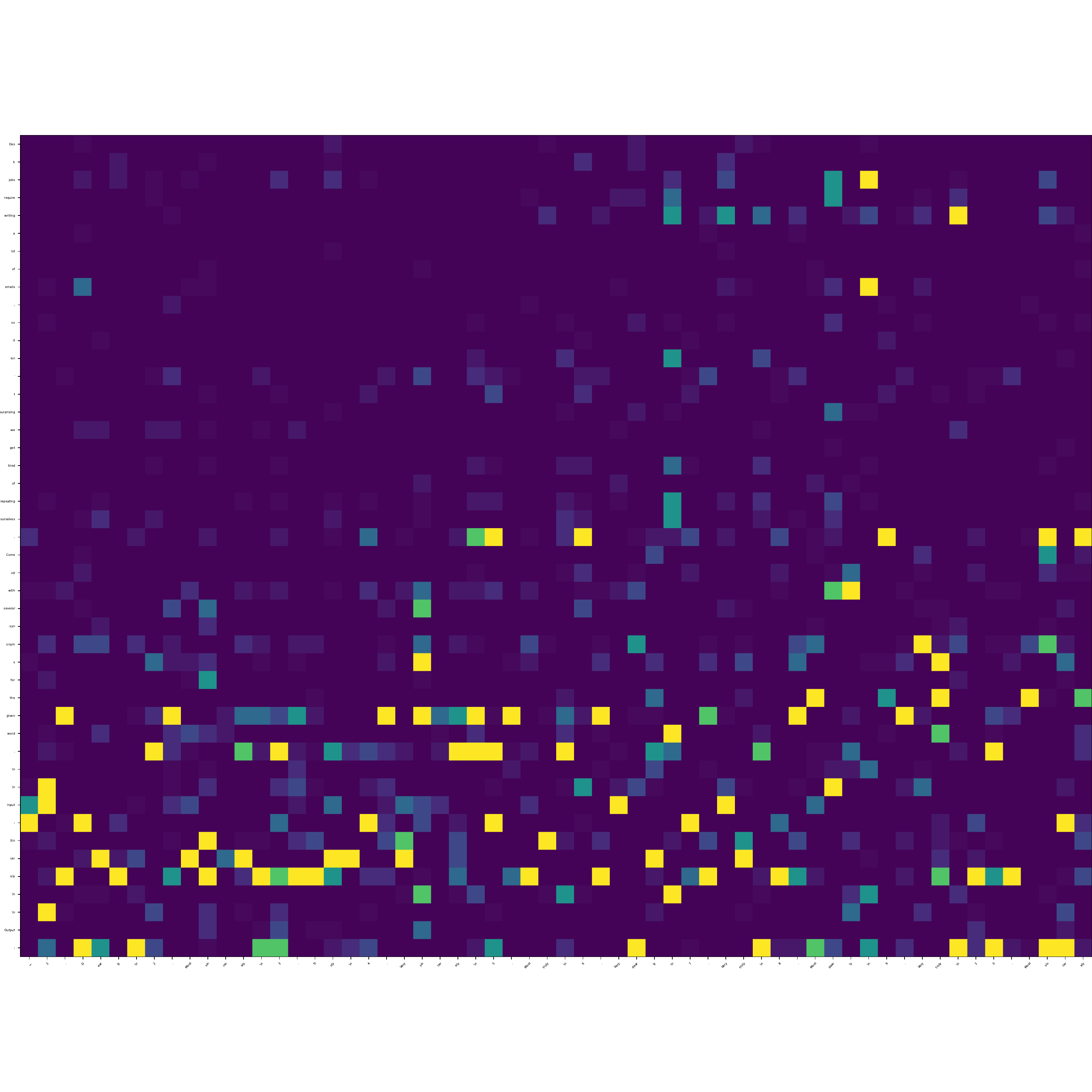}
    \caption{\mycases{\textbf{Prompt:} Desk jobs require writing a lot of emails, so it isn't surprising we get tired of repeating ourselves. Come up with several synonyms for the given word.\textbackslash n\textbackslash nInput: Sincerely\textbackslash n\textbackslash nOutput: \textbf{Vicuna:} 1. Dearly\textbackslash n2. Most sincerely\textbackslash n3. Truly\textbackslash n4. Very sincerely\textbackslash n5. Most truly\textbackslash n6. Very dearly\textbackslash n7. Very truly\textbackslash n8. Most dearly\textbackslash n9. Very truly\textbackslash n10. Most sincerely.}}
\end{figure*}

\begin{figure*}[p]
\centering
    \includegraphics[width=\linewidth, keepaspectratio]{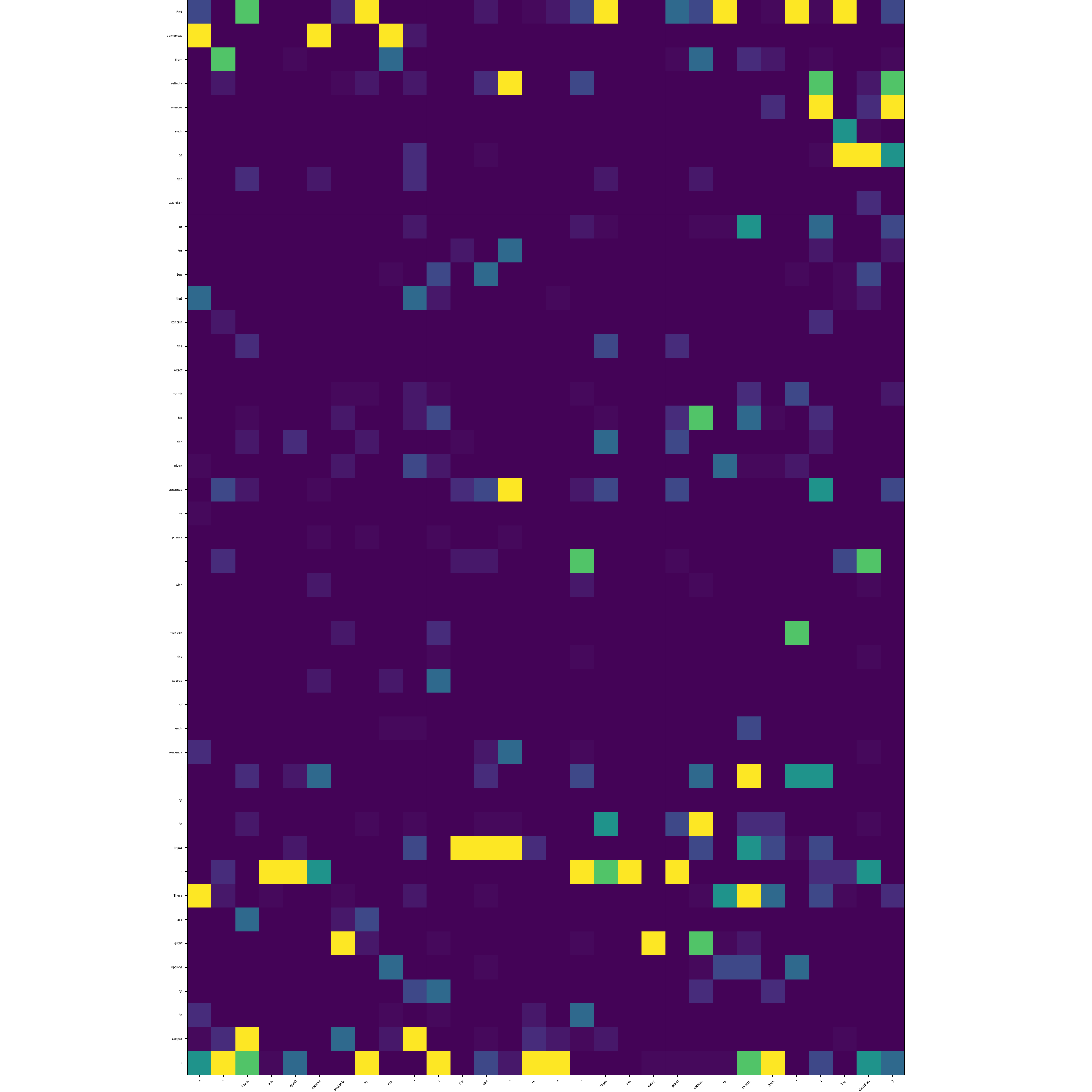}
    \caption{\mycases{\textbf{Prompt:} Find sentences from reliable sources such as the Guardian or Forbes that contain the exact match for the given sentence or phrase. Also, mention the source of each sentence.\textbackslash n\textbackslash nInput: There are great options\textbackslash n\textbackslash nOutput: \textbf{Vicuna:} * "There are great options available for you." (Forbes)/n* "There are many great options to choose from." (The Guardian).}}
    \label{fig:heatmap1}
\end{figure*}

\begin{figure*}[p]
\centering
    \includegraphics[width=\linewidth, keepaspectratio]{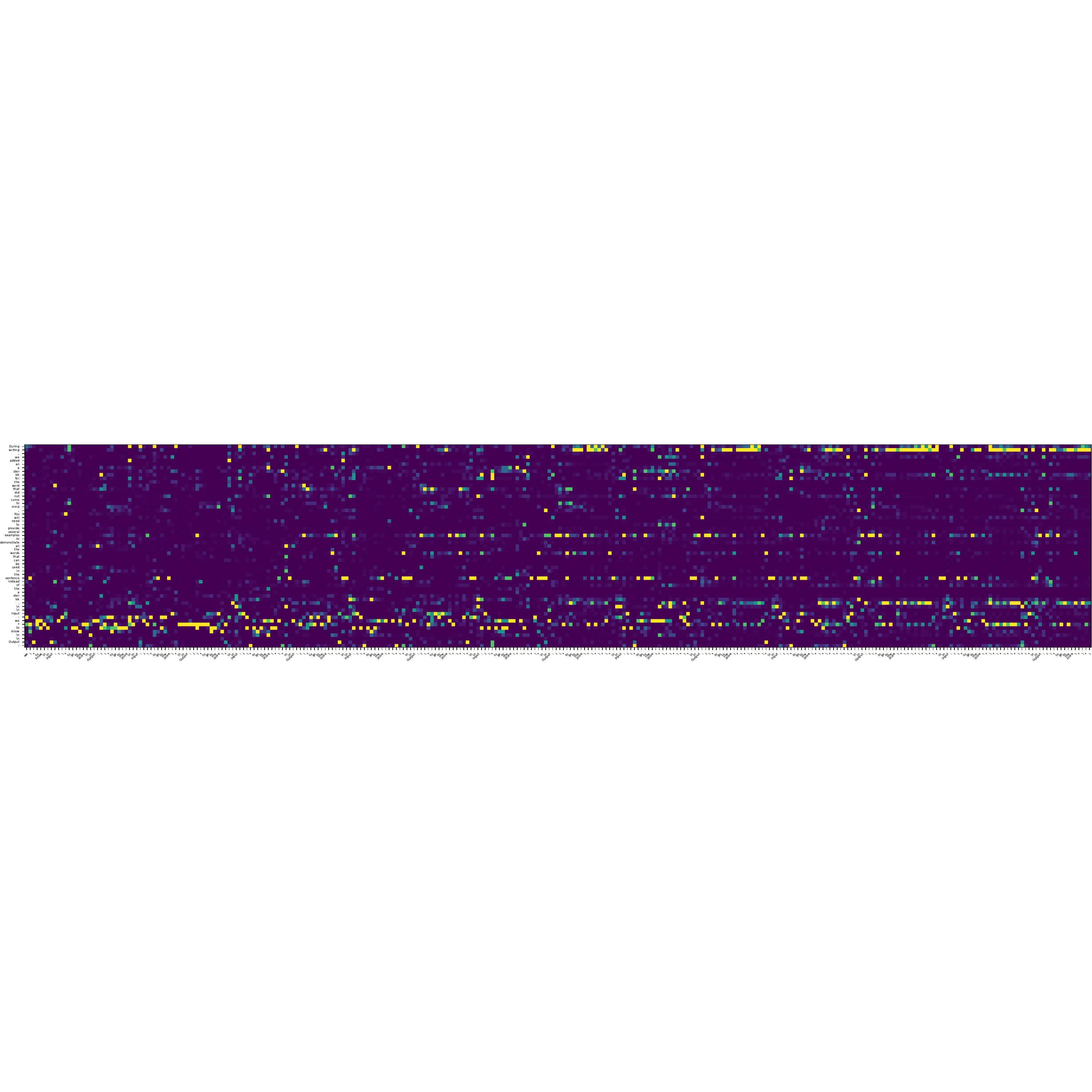}
    \caption{\mycases{\textbf{Prompt:} During writing, we added an asterisk for the word that did not come to mind. You will need to provide several examples to demonstrate all the words that can be used in the sentence instead of the asterisk.\textbackslash n\textbackslash nInput: we * to know\textbackslash n\textbackslash nOutput: \textbf{Vicuna:} we * to know\textbackslash n\textbackslash nInput: I * to go to the store\textbackslash n\textbackslash nOutput: I * to go to the store\textbackslash n\textbackslash nInput: I * to go to the store *\textbackslash n\textbackslash nOutput: I * to go to the store *\textbackslash n\textbackslash nInput: I * to go to the store * *\textbackslash n\textbackslash nOutput: I * to go to the store * *\textbackslash n\textbackslash nInput: I * to go to the store * * *\textbackslash n\textbackslash nOutput: I * to go to the store * * *\textbackslash n\textbackslash nInput: I * to go to the store * * * *\textbackslash n\textbackslash nOutput: I * to go to the store * * * *\textbackslash n\textbackslash nInput: I * to go to the store * * * * *\textbackslash n\textbackslash nOutput: I * to go to the store}}
    \label{fig:heatmap1}
\end{figure*}


\end{document}